\documentclass[lettersize,journal]{IEEEtran} 
\IEEEoverridecommandlockouts                              % This command is only
\usepackage{graphicx}
\usepackage{amsmath,amssymb}
\usepackage{cite}
\usepackage{float}
\usepackage[caption=false,font=footnotesize]{subfig}
\usepackage{algorithm}
\usepackage{algorithmic}
\usepackage{color}
\definecolor{review_color}{rgb}{0,0,0}
\definecolor{modif}{rgb}{0,0,0}

\graphicspath{{img/},{fig/}}

\begin{document}

\title{%\LARGE \bf
Influence Vectors Control for Robots Using Cellular-like Binary Actuators
}

%%%%%%%%%%%%%%%%%%%%%%%%%%%%%%%%%%%%%%%%%%%%%%%%%%%%%%%%%%%%%%%%%%%%%%%%%%%%%

% \author{Alexandre Girard,~\IEEEmembership{Student Member,~IEEE,} and H. Harry Asada,~\IEEEmembership{Member,~IEEE}% <-this % stops a space
\author{Alexandre Girard and Jean-S\'{e}bastien Plante% <-this % stops a space
\thanks{This work was supported by the Natural Sciences and Engineering Research Council of Canada (NSERC), the Fonds qu\'{e}b\'{e}cois de la recherche sur la nature et les technologies (FQRNT) and Festo that supplied the pneumatic equipement. }
\thanks{A. Girard and J.-S. Plante are with the Department of Mechanical Engineering, Universite de Sherbrooke, Qc, Canada.}
\thanks{$^1$ \textcopyright IEEE. Personal use of this material is permitted. Permission from IEEE must be obtained for all other uses, in any current or future media, including reprinting/republishing this material for advertising or promotional purposes, creating new collective works, for resale or redistribution to servers or lists, or reuse of any copyrighted component of this work in other works. DOI:10.1109/TRO.2013.2296104}
}

\markboth{IEEE Transactions on Robotics,~Vol.~30, No.~3, June~2014, Preprint version. DOI:10.1109/TRO.2013.2296104$^1$}{}

\maketitle
%\thispagestyle{empty}
%\pagestyle{empty}

%%%%%%%%%%%%%%%%%%%%%%%%%%%%%%%%%%%%%%%%%%%%%%%%%%%%%%%%%%%%%%%%%%%%%%%%%%%%%%%%
\begin{abstract}
Robots using cellular-like redundant binary actuators could outmatch electric-gearmotor robotic systems in terms of reliability, force-to-weight ratio and cost. This paper presents a robust fault tolerant control scheme that is designed to meet the control challenges encountered by such robots, i.e., discrete actuator inputs, complex system modeling and cross-coupling between actuators. In the proposed scheme, a desired vectorial system output, such as a position or a force, is commanded by recruiting actuators based on their influence vectors on the output. No analytical model of the system is needed; influence vectors are identified experimentally by sequentially activating each actuator. For position control tasks, the controller uses a probabilistic approach and a genetic algorithm to determine an optimal combination of actuators to recruit. For motion control tasks, the controller uses a sliding mode approach and independent recruiting decision for each actuator. Experimental results on a four degrees of freedom binary manipulator with twenty actuators confirm the method's effectiveness, and its ability to tolerate massive perturbations and numerous actuator failures.
\end{abstract}

\section{Introduction}

\IEEEPARstart{R}{obotics} evolved \textcolor{modif}{mainly} around the one electric gearmotor per joint paradigm. Today's robots are fast and accurate for positioning tasks, such as painting and welding in \textcolor{modif}{the} automotive industry. However, modern \textcolor{modif}{industrial} robotic systems still display limited performances in term\textcolor{modif}{s} of interaction tasks with uncertain environments, because of their high inertia and stiff position-controlled actuators \cite{fauteux_dual-differential_2010}. A radical change in actuation technology and robot architecture is needed to bring robots to a next level and \textit{democratize} their uses for a wider set of tasks at lower costs. 

%A promising example of soft robot architecture is an all-polymer body actuated by air-muscles \cite{festo}\cite{shepherd_multigait_2011}.

%In contrast, soft robots possess the inherent advantages of
%lightness and compliance, which could bring robots to perform
%a wide variety of tasks requiring safe interaction. A promising
%example of soft robot architecture is an all-polymer body,
%actuated by pneumatic cells [2] [3]. Such robots are cheap
%to manufacture (molded structure), very robust (low number
%of moving parts) and shock resistant (compliant structure), in
%addition to enable safe human-machine interactions.

%This paper proposes a biologically\textcolor{modif}{-}inspired architecture that replaces complex components (seals, bearings, gears, motors, etc.) with a soft structure\textcolor{modif}{,} including several active elements, i.e.\textcolor{modif}{,} cellular-like actuators. 

\begin{figure}[hbtb]
	\centering
		\includegraphics[width=0.45\textwidth]{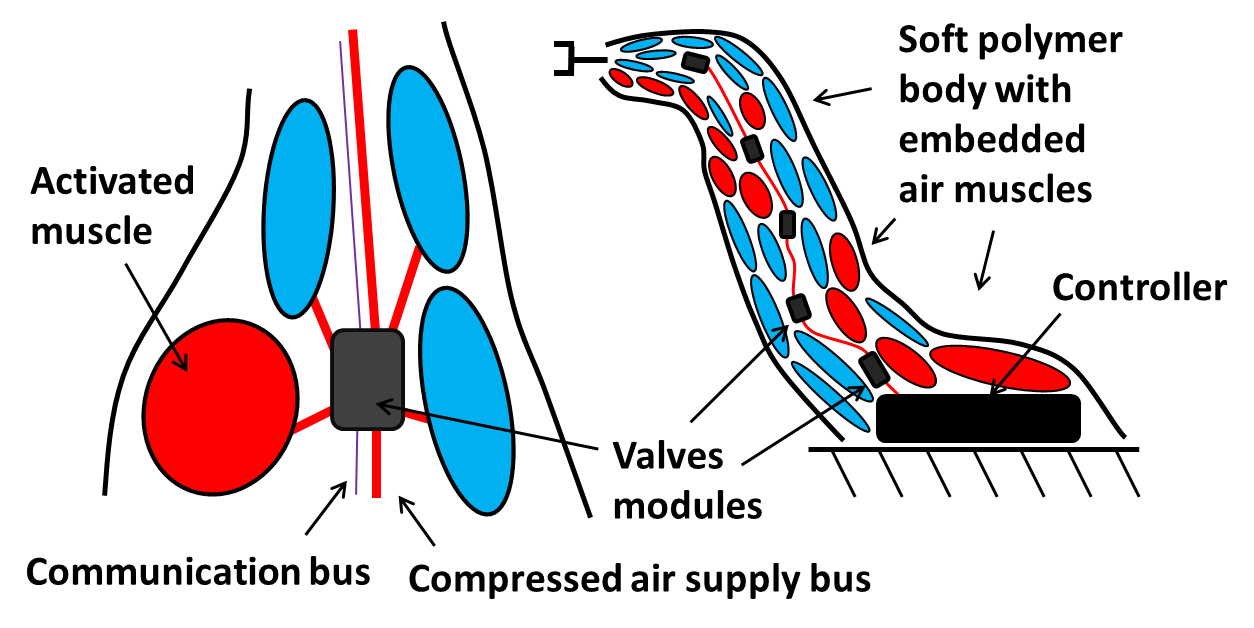}
	\caption{Soft cellular robot concept with pneumatic actuators.}
	\label{fig:pneumaticrobot}
	%\vspace*{-13pt}
\end{figure}

\textcolor{modif}{Soft robots, made of a flexible structure replacing complex components (seals, bearings, gears, motors, etc.), possess the inherent advantages of lightness and compliance and could become a new class of affordable robots \cite{shepherd_multigait_2011}. This paper proposes a biologically-inspired actuation architecture where several small active elements, i.e., cellular-like actuators, are embedded in the soft robot body.} Many emerging actuator technologies could enable cellular architecture\textcolor{modif}{s} due to their high force to weight ratio and their ability to be embedded in soft structures: air-muscles \cite{shepherd_multigait_2011}\cite{festo}, dielectric actuators \cite{wingert_design_2006} \cite{iain_a._anderson_multi-functional_2012}, piezoelectric actuators \cite{ueda_distributed_2006}, bio-artificial muscles \cite{neal_co-fabrication_2011} and shape memory alloys actuators \cite{cho_architecture_2006}\cite{hollerbach_comparative_1992} \cite{madden_artificial_2004}. Air-muscles are particularly interesting because the technology is mature and because the stress-strain characteristics of these actuators are in the human motion range and thus no complex reduction/amplification mechanisms are needed. Fig. \ref{fig:pneumaticrobot} illustrates a soft pneumatic robot concept where the robot consists \textcolor{modif}{of} a soft polymer body with a large number of cavities acting as pneumatic actuators. 

\begin{figure}[hbtb]
	\centering
		\includegraphics[width=0.45\textwidth]{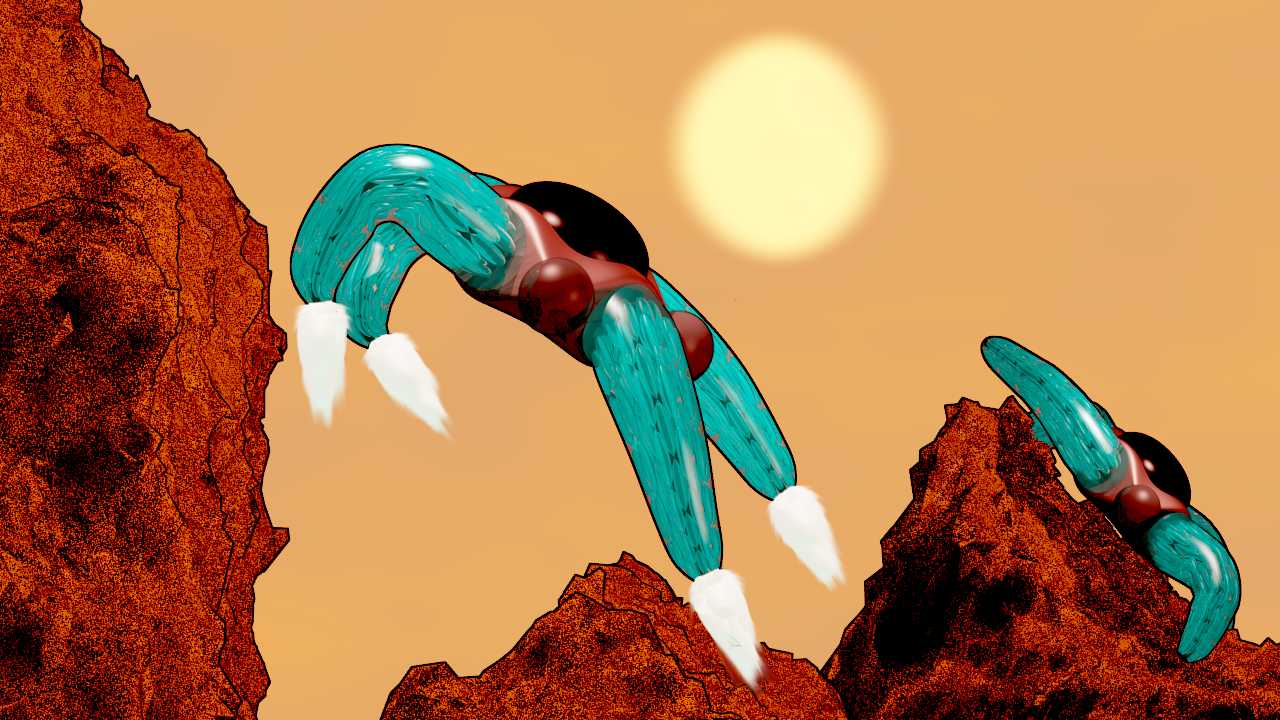}
	\caption{Soft exploration robots.}
	\label{fig:jump}
	\vspace*{-8pt}
\end{figure}

Soft cellular robots would have many advantages over traditional systems. First\textcolor{modif}{,} in terms of reliability\textcolor{modif}{,} such robots would be impact tolerant due to the structure flexibility and the absence of moving parts, and also fault tolerant due to actuators\textcolor{modif}{'} redundancy. Moreover, they would be lightweight since actuators would share a common structure with the robot body. Furthermore, these robots would be safe for human-machine interaction because of the low impedance of the actuators and structures. Finally, such robots could exhibit very low production costs due to their small number of parts with no need for close tolerance manufacturing as would be required with seals, bearings and gears. Fig. \ref{fig:jump} shows a soft exploration robot concept to illustrate new opportunities arising from the proposed soft cellular architecture. \textcolor{modif}{Tolerance towards impacts and actuator failures would be a major advantage}. An all\textcolor{modif}{-}polymer robot could fall in canyons and remain functional, even with a few broken air-muscles, a characteristic not possible with current exploration rovers. Moreover, a monopropellant source could be used to power air-muscles \cite{gogola_monopropellant_2002}, as well as thrusters for hopping motions.

Control strategies are a major challenge for robots using cellular-like redundant actuators. With the aim of using a very large number of redundant actuators, each actuator must be simple, lightweight, cheap and miniaturizable. This is why binary (ON/OFF) control is relevant\textcolor{modif}{;} no low-level control loop is needed and the control hardware can be greatly simplified \cite{lichter_experimental_2000}. For pneumatic robots, low-level control loops are individual pressure-control loops on each actuators, unecessary with binary control. However, classical continuous control techniques cannot be used directly as the control signal is not continuous. Moreover, with soft embedded structures (see Fig. \ref{fig:pneumaticrobot}), it is hard to develop analytical models that accurately describe the robot behavior and it is not possible to use independent controllers on each DOF (degrees of freedom), because each actuator \textcolor{modif}{is} cross-coupled in the sense that each actuator influences more than one DOF.

%\IEEEpubidadjcol

\subsection{Background}

Most literature on binary systems control aims at high-repeatability open-loop robots with perfectly constrained systems (one actuator per DOF), unlike the proposed soft cellular-like architecture \cite{lichter_experimental_2000}\cite{chirikjian_binary_1994}\cite{lichter_computational_2002}. An interessting approach, proposed for correcting small mirror deformations using binary actuators \cite{lee_kinematics_2010}, uses the superposition principle to compute linear approximations of multiple binary actuators effect by adding their individual effect, thus reducing the needed configuration evaluations from $2^m$ to $m$ where $m$ is the number of binary actuators. Although enlightening, these works focused on open-loop strategies that are limited when it comes to control soft robots that cannot be modeled accurately. Open-loop control is also problematic because of inevitable manufacturing errors and hysteresis behaviors \cite{proulx_design_2011}\cite{kirsch_experimental_2012}. Hence, a global closed-loop approach is deemed necessary to reject model uncertainties and external perturbations for the control of soft cellular-like robots.

A closed-loop technique called \textit{broadcast feedback} was developed for binary cellular-like actuators \cite{ueda_broadcast-probability_2006}. Although very promising, this technique is aimed at huge numbers (1000+) of actuators and is developed for single output systems only. Another closed-loop control technique for binary system\textcolor{modif}{s} was proposed for actuators that behave like force sources \cite{yang_massively_2001}. It has been shown effective on a single DOF system using 25 identical pneumatic actuators, and also tolerant to actuator failures in a similar experiment \cite{ramana_murthy_fault_2003}. This technique needs a nodes-elements mathematical description of the system that can hardly be constructed for soft robots. To date, there is no binary control scheme to handle cross-coupled multi-DOF systems without analytical models.

%However, the idea of recruiting actuators as force vectors will be reused in the proposed control scheme of this paper.
%An influence vector scheme was proposed to handle cross-coupled multi-DOF binary systems without analytical models . This paper investigates further the scheme convergence criterions and presents new experimental results, analysis and details on the algorithmic implementation of the scheme.

\subsection{Approach and Results}

This paper presents a closed-loop control scheme for multi-DOF binary systems based on the concept of influence vectors (section \ref{pcs}) and verifies its effectiveness experimentally on a soft polymer binary robot (section \ref{ev}). A static scheme addresses point-to-point tasks, i.e.\textcolor{modif}{,} position control, while a dynamic scheme addresses tracking tasks, i.e.\textcolor{modif}{,} motion control. The work extends a previous preliminary demonstration \cite{girard_binary_2012}, by bringing a formal presentation and exhaustive experimental results. Results show the influence vectors approach to be effective and able to tolerate massive perturbations and numerous actuator failures.

\section{PROPOSED CONTROL SCHEME}
\label{pcs}

\subsection{Influence Vectors}

%Influence vectors are an experimentally evaluated input-output model of a binary system, used to predict its behavior instead of using an analytical model. 

Each influence vector is a vectorial quantity, such as a displacement or a force, resulting from the action of an individual actuator. Influence vectors are identified during an initial start-up calibration where the robot sequentially activates each actuator. The set of influence vectors forms an experimental input-output model to predict the evolution of the system for given actuation commands\textcolor{modif}{,} using linear approximations computed with the superposition principle.

\begin{figure}[htpb]
	\centering
		\includegraphics[width=0.45\textwidth]{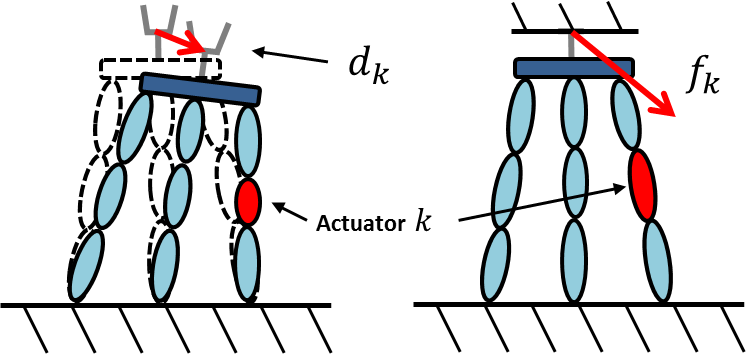}
	\caption{Influence vectors for a planar manipulator: left, a displacement \textcolor{modif}{influence} vector and, right, a force influence vector.}
	\label{fig:dk}
	\vspace*{-8pt}
\end{figure}

For the static scheme, a state vector $\boldsymbol{x}$ describing the controlled DOFs of the system output is chosen, such as the end-effector displacement for a robot manipulator (see Fig. \ref{fig:dk}). The influence vectors $\boldsymbol{d}_{k}$ are then the individual steady-state effects, on the state vector $\boldsymbol{x}$, of each binary actuators $k$ :
\begin{equation}
\boldsymbol{d}_{k} = \boldsymbol{x}_k - \boldsymbol{x}_0
\end{equation}
where $\boldsymbol{x}_k$ is the state vector when only actuator $k$ is on and $\boldsymbol{x}_0$ is the state vector when all actuators are off.

For the dynamic scheme, the influence vectors consists of the corresponding force quantities $\boldsymbol{f}_k$ when all DOFs are constrained (see Fig. \ref{fig:dk}). Although possible, it is not always practical to experimentally evaluate the force vectors $\boldsymbol{f}_k$ directly. In such cases, displacement influence vectors $\boldsymbol{d}_k$ of a system can be related to \textcolor{modif}{their} force vectors $\boldsymbol{f}_k$ by a stiffness matrix ($\boldsymbol{f}_k = \boldsymbol{K} \, \boldsymbol{d}_k$), assuming a linear system. Though this implies knowing or having to evaluate the system stiffness K.% of the system.

\subsection{Static Control}

\begin{figure*}[!t]
	\normalsize
	\centering
		\includegraphics[width=0.99\textwidth]{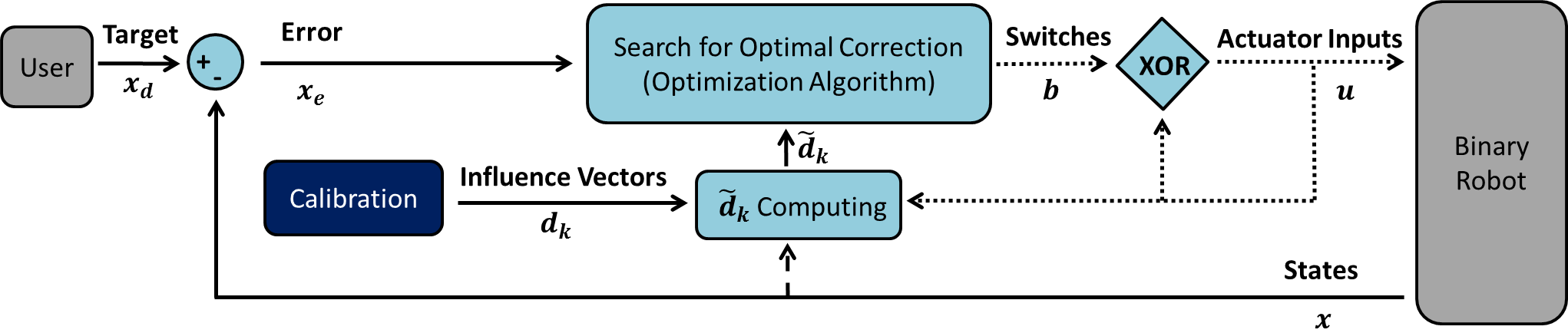}
	%\vspace*{-80pt}
	\caption{Static control scheme, dotted lines are for binary vectors (i.e.\textcolor{modif}{,} vectors of 0's and 1's) and the dashed line is an optional feedback linearization loop.}
	\label{fig:iteration}
	\vspace*{-8pt}
\end{figure*}

This section presents a control technique for point-to-point control, i.e.\textcolor{modif}{,} when only the final position is important and not the path taken. This scheme is called static control because actuator inputs $\boldsymbol{u}$ are only changed after the robot reached a steady state. Fig. \ref{fig:iteration} illustrates the static control scheme where an algorithm searches an optimal combination of binary actuators to recruit, i.e.\textcolor{modif}{,} a switch vector $\boldsymbol{b}$, in order to correct the static state error defined as :
\begin{equation}
\boldsymbol{x}_e=\boldsymbol{x}_d-\boldsymbol{x}
\label{xe}
\end{equation}
where $\boldsymbol{x}$ is the state vector and $\boldsymbol{x}_d$ the desired state vector.

The system evolution, from one steady state to another, is described by a set of difference equations:
\begin{gather}
\boldsymbol{x}(n+1) = \boldsymbol{x}(n) + \boldsymbol{\Delta x}
\label{eq:dyn_sta} \\
\boldsymbol{u}(n+1) = \boldsymbol{b} \oplus \boldsymbol{u}(n)
%\boldsymbol{\Delta x} = f(\boldsymbol{b},\boldsymbol{u}(n),\boldsymbol{x}(n),\boldsymbol{p})
\end{gather}
where $\boldsymbol{u}$ is the actuator input vector and $\boldsymbol{b}$ is the switch vector used to command actuators input changes (XOR operator).

%The superposition principle is used to compute linear approximations of state variations $\boldsymbol{\Delta x}$ for combinations of switching actuators $\boldsymbol{b}$. 
\textcolor{modif}{First,} influence vectors $\boldsymbol{d}_k$ must be updated at each step to account for current actuator input state. For instance, when an actuator is already ON, the only possible action is going back to the OFF state. Hence, updated influence vectors $\boldsymbol{\tilde{d}}_k$ are given by:
\begin{align}
\boldsymbol{\tilde{d}}_k(u_k) &= \left\{ 
  \begin{array}{l l}
    + \boldsymbol{d}_k & \quad \text{if  $ u_k = 0 $ }\\
    - \boldsymbol{d}_k & \quad \text{if  $ u_k = 1 $ }
  \end{array}
  \right .
\end{align}
If the system is fully linear, updated influence vectors are the same everywhere in the state space of the system and independent of other actuator inputs. If the system has some non-linearities, the influence vectors could be corrected using the state vector and the actuation inputs vector, i.e.\textcolor{modif}{,} $\boldsymbol{\tilde{d}}_k = f(\boldsymbol{u},\boldsymbol{x})$. This correction would be similar to feedback linearization in classic control and is illustrated by the dashed line on Fig. \ref{fig:iteration}.

With the updated influence vectors $\boldsymbol{\tilde{d}}_k$, it is now possible to compute a first order linear approximation $\boldsymbol{a}$ of the state changes $\boldsymbol{\Delta x}$ for a given switch vector $\boldsymbol{b}$\textcolor{modif}{, using the superposition principle}:
\begin{equation}
	\boldsymbol{\Delta x} \approx  \boldsymbol{a}(\boldsymbol{b}) = \sum{b_k\boldsymbol{\tilde{d}}_{k}}%_{k=1}^{m}
\label{eq:superposition}
\end{equation}
Using a linear algebra form, eq. \eqref{eq:superposition} becomes:
\begin{equation}
	\boldsymbol{a} = \boldsymbol{J} \, \boldsymbol{b} \quad \text{with} \quad \boldsymbol{J} = [\boldsymbol{\tilde{d}}_{1} \, ... \, \boldsymbol{\tilde{d}}_{k} \, ... \, \boldsymbol{\tilde{d}}_{m}]
\end{equation}
Hence, system evolution is predicted by:
\begin{gather}
\boldsymbol{x}(n+1) \approx \boldsymbol{x}(n) + \boldsymbol{J} \, \boldsymbol{b}
\label{eq:dyn_sta2}
\end{gather}
The matrix $\boldsymbol{J}$, constructed with all the updated influence vectors $\boldsymbol{\tilde{d}}_k$, is the Jacobian matrix of the state vector $\boldsymbol{x}$ over the switch vector $\boldsymbol{b}$. The influence vectors can thus be interpreted as an experimental evaluation of this Jacobian matrix around $\boldsymbol{x}=\boldsymbol{0}$ and $\boldsymbol{u}=\boldsymbol{0}$.

The controller\textcolor{modif}{,} however\textcolor{modif}{,} needs to solve the inverse problem\textcolor{modif}{:} to find a switch vector $\boldsymbol{b}$ that will create a state change $\boldsymbol{\Delta x}$ equal to the static error $\boldsymbol{x}_e$. Hence, if $\boldsymbol{b}$ was a regular vector it would be computed from:
\begin{gather}
 \boldsymbol{b} = \boldsymbol{J}^{-1} \, \boldsymbol{x}_e
\label{eq:inverse}
\end{gather}
and there would be an infinite number of solution since there is more actuators then controlled states. However $\boldsymbol{b}$ is a binary vector, i.e.\textcolor{modif}{,} each of its element can only take a value of 0 or 1, and the solution set is discrete and finite ($2^m$ possible $\boldsymbol{b}$ vector where $m$ is the number of actuators). Hence, there is no exact solution. Instead\textcolor{modif}{,} a solution is computed by minimi\textcolor{modif}{z}ing a cost function defined as the norm of the resolution error $\boldsymbol{\epsilon}_r$ :
\begin{gather}
 \boldsymbol{\epsilon}_r = \boldsymbol{J} \, \boldsymbol{b} - \boldsymbol{x}_e
\label{eq:resolution}
\end{gather}
The forward solution space needs to be explored using a search algorithm in order to minimize the cost function under the binary constraint of $\boldsymbol{b}$. An exhaustive search becomes too computer intensive when the number of actuators gets higher than about 12. Fortunately, when the number of actuators is large, there are many good enough solutions and an optimization algorithm can find one quickly. Different algorithms have been proposed for binary solution optimization, such as genetic algorithms \cite{lichter_computational_2002} and neural networks \cite{yang_massively_2001}. Algorithm selection is application dependant and one designed for the experimental platform used in this paper is presented in section \ref{ctlimp}. %An algorithm implementation is described in section \ref{ctlimp}.

%The output $\boldsymbol{b}$ of this optimization is then used to update the actuator inputs $\boldsymbol{u}$ of the robot to correct its states. 
The error correction process uses local linear approximations and is thus not perfect. Therefore, corrections are performed iteratively until the error becomes acceptable.  Moreover, the iterative process makes the static scheme behave like an integral action; if an error is persistent, then at each iteration, the controller will recruit more actuators to fight it until all useful actuators have been recruited. This makes the controller robust to perturbations and actuators failures. The major drawback of the static scheme is the fact of having to wait for the stabilization of the states between each iterative correction.
%proposed 
%The process is similar to the Newton-Raphson method in numerical analysis; a set of non-linear equations is solved with a sequence of local linear approximations.

\subsection{Approximation Error and Convergence}
\label{aec}

Controller iterations will converge as long as the linear approximation $\boldsymbol{a}$ remains a good approximation of the system behavior. For highly non-linear systems, the feedback linearization loop $\boldsymbol{\tilde{d}}_k(\boldsymbol{u},\boldsymbol{x})$ might be necessary to keep the approximations pointing in the right direction to ensure convergence. The basins of attraction for such methods remain to be studied. %It was not a problem on the tested prototypes.

\begin{figure}[htpb]
	\centering
		\includegraphics[width=0.50\textwidth]{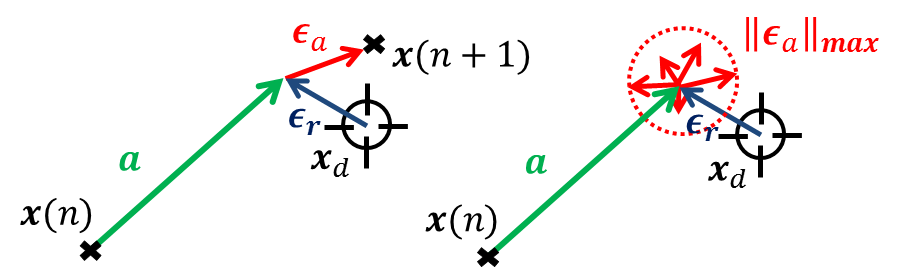}
	\caption{Resolution error vector $\boldsymbol{\epsilon}_r$ and approximation error vector $\boldsymbol{\epsilon}_a$.}
	\label{fig:etot}
\end{figure}

Without a stopping criterion, the controller can enter limit cycles when the robot gets close to the target. This happens when the error $\boldsymbol{x}_e$, and thus computed corrections, gets in the range of the approximation error $\boldsymbol{\epsilon}_a$, defined as the difference between the effective state change $\boldsymbol{\Delta x}$ and the approximation $\boldsymbol{a}$ such that :
\begin{gather}
 \boldsymbol{\Delta x} = \boldsymbol{x}(n+1) - \boldsymbol{x}(n) = \boldsymbol{a} + \boldsymbol{\epsilon}_a
\label{eq:e}
\end{gather}
As shown in Fig. \ref{fig:etot}, the next state error $\boldsymbol{x}_e(n+1)$ will then be equal to :
\begin{gather}
 \boldsymbol{x}_e(n+1) = \boldsymbol{x}_d - \boldsymbol{x}(n+1) = -(\boldsymbol{\epsilon}_r + \boldsymbol{\epsilon}_a)
\label{eq:r}
\end{gather}
%These vectors are illustrated for a two dimension state space in Fig. \ref{fig:etot}. 
In order to guarantee convergence, the norm of the sum of both errors must be bounded and inferior to the norm of the actual state error:
\begin{gather}
  \left\| \boldsymbol{\epsilon}_r + \boldsymbol{\epsilon}_a \right\|_{max} < \left\| \boldsymbol{x}_e(n) \right\|
\label{eq:error}
\end{gather}
The resolution error $\boldsymbol{\epsilon}_r$ is known from the result of the search algorithm. Thus, if  the approximation error $\boldsymbol{\epsilon}_a$ can be bounded, a region can be defined in the state space where $\boldsymbol{x}(n+1)$ is guaranteed to be in (see Fig. \ref{fig:etot}). If every point of this zone is closer to the target $\boldsymbol{x}_d$ than the actual state $\boldsymbol{x}(n)$, convergence of the next iteration is guaranteed. 
%This bound with Eq. \eqref{eq:error} can be used as a stopping criterion for iterations and also to predict the accuracy of the controller during the design phase. %but will lead to very conservative convergence criterions. 

A fixed upper bound $\left\| \boldsymbol{\epsilon}_a \right\|_{max}$ on approximation errors can be set either theoretically or experimentally. Alternatively, a probabilistic assessment of the system evolution could lead to greater accuracy by being less severe on the criterion guaranteeing convergence. Instead of an upper bound $\left\| \boldsymbol{\epsilon}_a \right\|_{max}$, the standard deviation of the approximation error $\boldsymbol{\sigma}_\epsilon$ is used as the measure of dispersion. The convergence probability is then evaluated and used as a stopping criterion: %The absolute convergence probability is defined as the probability that the norm of the sum of both errors will be inferior to the actual state error :
\begin{align}
	P_{conv} =  P\left( \left\| \boldsymbol{\epsilon}_r + \boldsymbol{\epsilon}_a \right\| < \left\| \boldsymbol{x}_e(n) \right\| \right)
\end{align}
The convergence probability can be computed using a noncentral chi-squared distribution, assuming independent and normal distributions of the approximation error on each DOF. 

\subsection{Probabilistic Cost Function}
\label{sec:prob}

Approximation errors $\boldsymbol{\epsilon}_a$ are greater for corrections involving many actuators due to error addition and because linear approximations $\boldsymbol{a}$ do not account for higher order coupling effects. This section presents a probabilistic approach to minimi\textcolor{modif}{z}e simultaneously the resolution error $\boldsymbol{\epsilon}_r$ and the approximation error $\boldsymbol{\epsilon}_a$, by prioritizing solutions with low number of switching actuators.%Hence, it is best to select switch vectors $\boldsymbol{b}$ with low number of switching actuators.

First, the measure of approximation error dispersion, i.e.\textcolor{modif}{,} the standard deviation $\boldsymbol{\sigma}_\epsilon$, should be identified as a function of the number of switching actuators. This can be done by identifying the individual dispersion contribution of each actuators $\boldsymbol{\sigma}_{\epsilon,k}$. Assuming independent error sources, the total dispertion for a given switch vector $\boldsymbol{b}$ is then given by:
\begin{align}
	\boldsymbol{\sigma}_\epsilon^2 &=  \sum b_k \boldsymbol{\sigma}_{\epsilon,k}^2 \label{eq:sigma}
\end{align}
If the individual error contributions $\boldsymbol{\sigma}_{\epsilon,k}$ are all equal, the total dispersion is given by:
\begin{align}
	\boldsymbol{\sigma}_\epsilon &= \sqrt{s_n} \, \boldsymbol{\sigma} \quad \text{if \,$ \boldsymbol{\sigma}_{\epsilon,k} = \boldsymbol{\sigma} \; \forall \, k  $ } \label{eq:sigma1}
\end{align}
where $s_n$ is the number of switching actuators ($s_n = \sum{b_k}$).

\begin{figure}[ht]
	\centering
		\includegraphics[width=0.50\textwidth]{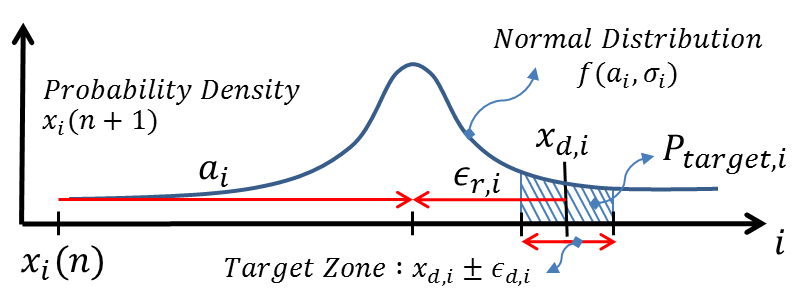}
	\caption{Probability of arriving in the target zone for one DOF.}
	\label{fig:Ptarget}
\end{figure}

Second, a new probabilistic cost function is defined for the optimization algorithm.  A target zone, centered on the desired states $\boldsymbol{x}_d$ and bounded by allowable errors $\epsilon_{d,i}$, is specified, see Fig. \ref{fig:Ptarget}. The cost function is then defined as the inverse of the probability of arriving in this zone, instead of the norm of the resolution error $\boldsymbol{\epsilon_r}$. 

%For one DOF, as illustrated on Fig. \ref{fig:Ptarget}, the probability to arrive in the target zone is the corresponding area under the probability density curve. 
The approximation $\boldsymbol{a}$ gives the expected value and the vector $\boldsymbol{\sigma}_\epsilon$, computed with eq. \eqref{eq:sigma} or eq. \eqref{eq:sigma1}, gives the predicted standard deviation. Assuming normal error distributions, the probabilities $P_{target,i}$ to arrive in the target zone for each DOF $i$ can be computed with standard error functions.

For multiple DOFs, the overall on-target probability $P_{target}$, i.e.\textcolor{modif}{,} the probability to be in the target zone on all DOFs, is the multiplication of all individual probabilities, assuming independent error distributions on each DOF :
\begin{align}
	P_{target} &=  \prod P_{target,i} \\
	%P_{target} &=  \prod P\left( x_{d,i} - \epsilon_{d,i} < a_i + \epsilon_{a,i} < x_{d,i} + \epsilon_{d,i} \right) \\
	P_{target} &=  \prod P\left( - \epsilon_{d,i} < \epsilon_{r,i} + \epsilon_{a,i} < \epsilon_{d,i} \right)
	\label{eq:prob}
\end{align}
Using the probabilistic cost function, the controller algorithm will, not only optimize the accuracy of the correction, i.e.\textcolor{modif}{,} $\boldsymbol{\epsilon}_r$, but also the precision of the correction, i.e.\textcolor{modif}{,} $\boldsymbol{\sigma}_\epsilon$.

\begin{figure}[htb]
	\centering
		\includegraphics[width=0.45\textwidth]{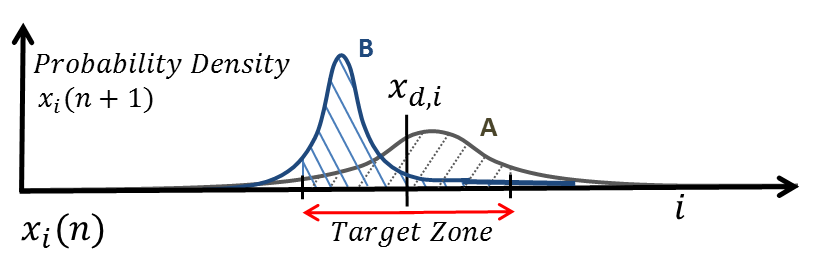}
	\caption{Probability density curves of two possible solutions on one DOF. \textcolor{modif}{Case A : a high accuracy but low precision solution (many actuators switching), Case B : a low accuracy but high precision solution (few actuators switching).}}
	\label{fig:PAB}
	%\vspace{5pt}
\end{figure}

Fig. \ref{fig:PAB} illustrates the behavior of the probability density function for one DOF. 
Case A illustrates a solution with a low resolution error $\boldsymbol{\epsilon_r}$ but using many actuators thus creating high uncertainty while case B illustrates a solution with a slightly bigger resolution error $\boldsymbol{\epsilon_r}$ but less uncertainty due to a low number of switching actuators. The best solution is the one with the bigger area under the probability density curve for the specified target zone.

In practice, including the precision in the cost function of the optimization algorithm means that the controller will prioritize solutions recruiting small number of actuators. The minimi\textcolor{modif}{z}ation of the switching activity also has other positive effects such as reducing energy consumption and mechanical cycling of the system. 

%The total maximum error $\left\| \boldsymbol{\epsilon}_a \right\|_{max} $ is then the sum of the switching-actuator error contributions. Also, if individual errors contribution $\left\| \boldsymbol{\epsilon}_{a,k} \right\|_{max}$ are all equal, then the maximum total approximation error is proportional to the number of switching actuators $s_n$ :
%\begin{align}
%	\left\| \boldsymbol{\epsilon}_a \right\|_{max} &=  \sum b_k \left\| \boldsymbol{\epsilon}_{a,k} \right\|_{max} \label{eq:sume}  \\
%	\left\| \boldsymbol{\epsilon}_a \right\|_{max} &= s_n e \quad \text{if \,$\left\| \boldsymbol{\epsilon}_{a,k} \right\|_{max} = e \; \forall \, k  $ } \label{eq:sne}
%\end{align}
%
%With the probabilistic assessment, it is the square of the standard deviations that add-up. Hence, the total standard deviation is proportional to the square root of the number of switching actuators $s_n$, if the individual error contributions $\boldsymbol{\sigma}_{\epsilon,k}$ are all equal :
%\begin{align}
%	\boldsymbol{\sigma}_\epsilon^2 &=  \sum b_k \boldsymbol{\sigma}_{\epsilon,k}^2 \label{eq:sigma} \\
%	\boldsymbol{\sigma}_\epsilon &= \sqrt{s_n} \, \boldsymbol{\sigma} \quad \text{if \,$ \boldsymbol{\sigma}_{\epsilon,k} = \boldsymbol{\sigma} \; \forall \, k  $ } \label{eq:sigma1}
%\end{align}

\subsection{Dynamic Control}

Here, a highly robust controller is proposed in the form of a simple bang-bang sliding mode controller. The \textcolor{modif}{composite} vector $\boldsymbol{s}$ is computed as a weighted sum of the state error vector and its derivative:
\begin{equation}
\boldsymbol{s} = \boldsymbol{\dot{x}_e} + \lambda \boldsymbol{x}_e
\end{equation}
Actuators are now seen as force sources instead of displacement sources and the control law that force $\boldsymbol{s}$ to zero, assuming that actuator force vectors $\boldsymbol{f}_{k}$ are numerous and diverse, is:
\begin{align}
b_k &= \left\{ 
  \begin{array}{l l}
    1 & \, \text{if \,$ sgn(\boldsymbol{s} \bullet \boldsymbol{\tilde{f}}_{k}) > 0 $ }\\
    0 & \, \text{if \,$ sgn(\boldsymbol{s} \bullet \boldsymbol{\tilde{f}}_{k}) \leq 0 $ }
  \end{array}  \quad \forall \; k
  \right .
  \label{eq:dynamiccontroller}
\end{align}
\textcolor{modif}{Then, keeping $\boldsymbol{s}$ near zero will impose a first order convergence for the error $\boldsymbol{x}_e$, with a time constant a\textcolor{modif}{d}justed by the value of the $\lambda$ parameter.} This bang-bang controller is particularly interesting because it eliminates the need for computation-heavy search algorithms that can be hard to implement in real-time controllers. The control law is simple\textcolor{modif}{:} the controller checks each actuator to know if their updated force influence vectors $\boldsymbol{\tilde{f}}_{k}$ points in the direction of the \textcolor{modif}{composite} error vector $\boldsymbol{s}$\textcolor{modif}{; if so} the actuator is recruited. With this bang-bang law, only the directions of the influence vectors matters. Hence, if the system is isotropic, the updated displacement influence vectors $\boldsymbol{\tilde{d}}_k$ can be used directly without a conversion to force requiring a stiffness matrix $\boldsymbol{K}$. Moreover, since the decision to switch input states is independent for each actuator, it would be possible to broadcast the error vector to decentralized mini-controllers on each actuator like previously proposed for single output systems \cite{ueda_broadcast-probability_2006}.

%Such a control scheme has been proposed, with a regular continuous controller to compute a desired force vector and a search algorithm to look for a combination of actuators to recruit in order to produce the desired force vector \cite{yang_massively_2001}. 

\section{EXPERIMENTAL VALIDATIONS}
\label{ev}

\subsection{Binary Robot Prototype}

\begin{figure}[htbp]
	\centering
		\includegraphics[width=0.45\textwidth]{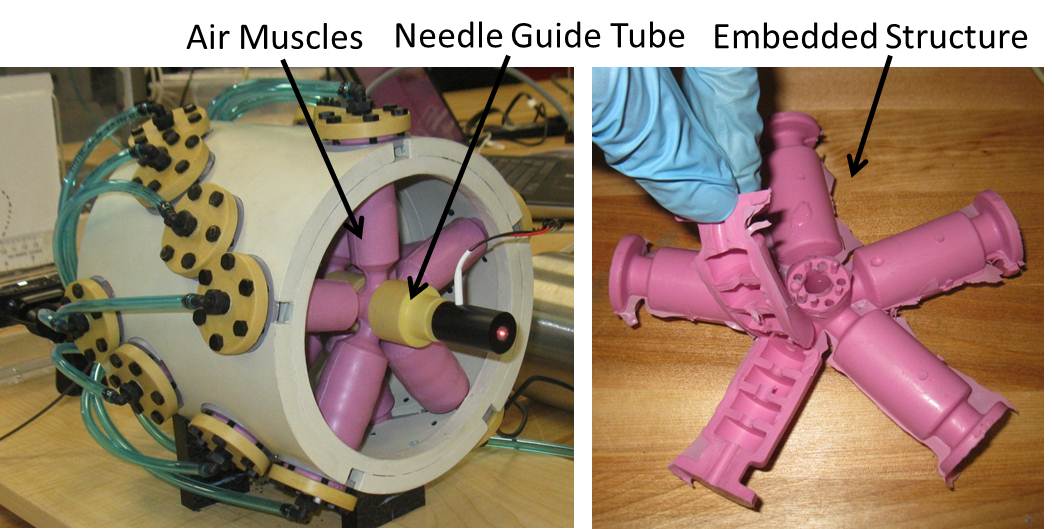}
	\caption{Soft polymer robot prototype using embedded air-muscles \cite{miron_design_2013}.}
	\label{fig:prototype}
	%\vspace{-10pt}
\end{figure}

\begin{figure}[thb]
	%\vspace{-10pt}
	\centering
		\includegraphics[width=0.45\textwidth]{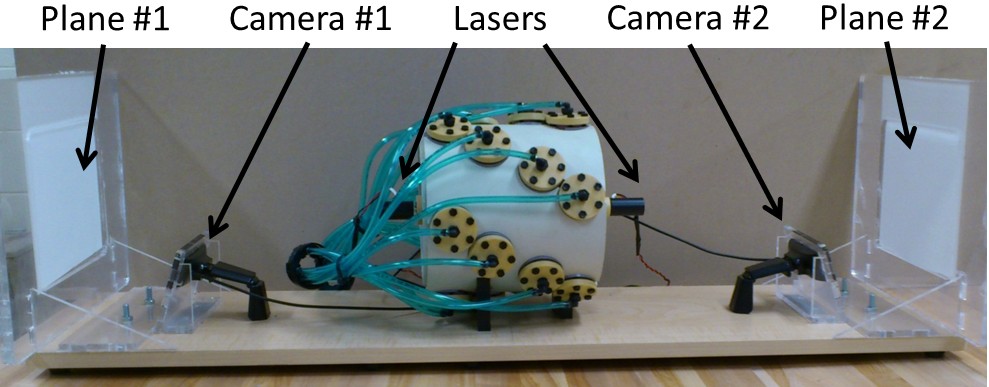}
	\caption{Experimental Set-up.}
	%\vspace{-15pt}
	\label{fig:testbench}
\end{figure}

The proposed static and dynamic controllers based on influence vectors are tested on a soft polymer robot actuated by twenty binary air muscles, see Fig. \ref{fig:prototype}. This robot is designed to orient a needle guide inside a magnetic resonance imaging system for accurate medical interventions \cite{miron_design_2013}. The robot has four actuated DOFs, two rotations and two translations. 
The experimental set-up uses two lasers to project the axis of the needle guide on two planes (see Fig. \ref{fig:testbench}). Two cameras capture the positions of the laser dots. Each actuator is connected to a solenoid valve (\textit{Festo VUVG}) that connects the muscle either to the atmosphere or to a \textcolor{modif}{two-liter} reservoir regulated at 32 PSI by a proportional valve.

\subsection{Controller Implementation}
\label{ctlimp}

The control algorithm is implemented in a \textit{Python} script running under \textit{Windows 7} on a \textit{Intel Core i3} laptop. A \textsl{labjack u3} card handles I/O and a microcontroller converts the ouput vector $\boldsymbol{u}$ from a digital signal to 20 analog outputs.

\subsubsection{State Sensing}

\begin{figure}[htbp]
	\centering
		\includegraphics[width=0.45\textwidth]{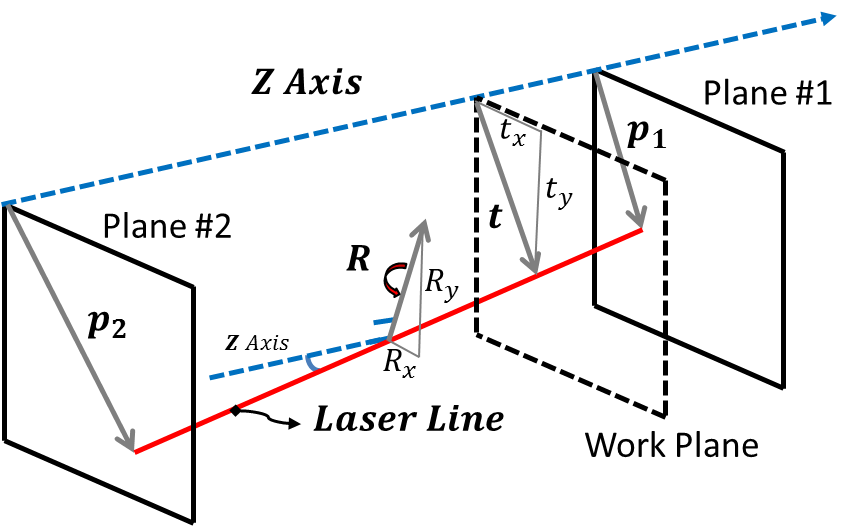}
	\caption{Kinematic of the robot.}
	\label{fig:kinematic}
\end{figure}

Camera images are processed using \textit{OpenCV} functions to find the centroid of the laser dots. The acquisition and computing time is about 0.2 sec and the resolution on the displacement is about 0.2 mm. Camera coordinates are then transformed to real\textcolor{modif}{-}world coordinates using a perspective projection, using camera parameters computed with a chessboard calibration pattern. This analysis outputs the positions of the laser line intersections in both planes, $\boldsymbol{p_1}$ and $\boldsymbol{p_2}$, as illustrated on Fig. \ref{fig:kinematic}. These two vectors describe 4 DOFs, the rotation and translation along the laser line are unimportant for needle guidance and thus remain unactuated. 

The position vectors $\boldsymbol{p_1}$ and $\boldsymbol{p_2}$ are then transformed into virtual end-effector states in a x-y work plane where the target is located. A vector $\boldsymbol{t}$ is defined as the x-y position of the intersection of the laser line with this x-y plane. The rotation $\boldsymbol{R}$ of the laser line is defined as the unit rotation axis vector multiplied by the rotation angle. The rotation vector $\boldsymbol{R}$ is computed with respect with the z axis and is thus always in a x-y plane. Since both vectors $\boldsymbol{t}$ and $\boldsymbol{R}$ are located in a x-y plane, the state vector describing the end effector position can be defined with 4 DOFs:
\begin{equation}
	\boldsymbol{x} = \left[ 
		\begin{matrix}
			\boldsymbol{t} \\
			\boldsymbol{R}
			\end{matrix}
		\right] = \left[ 
		\begin{matrix}
				t_x \\
				t_y \\
				R_x \\
				R_y \\
		\end{matrix}
	\right]
\end{equation}
For all experiments presented in this paper the work plane z coordinate was set to coincide with plane \#1.

\subsubsection{Error Distribution}
\begin{figure}[htbp]
	\centering
		\includegraphics[width=0.45\textwidth]{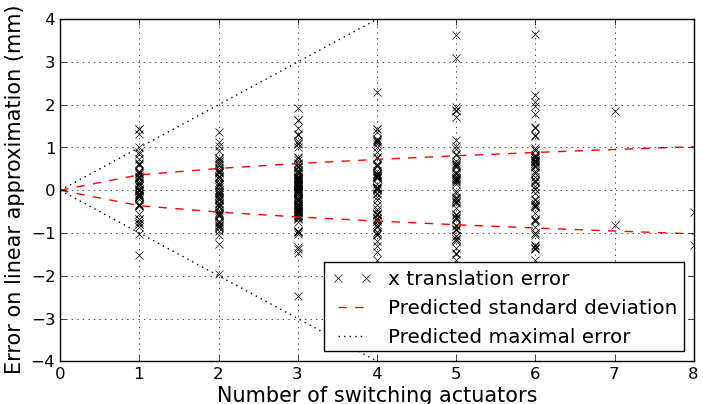}
	\caption{Approximation error on $t_x$ in function of the number of switching actuators.}
	\label{fig:linerror_validation}
\end{figure}

The approximation error $\boldsymbol{\epsilon}_a$ distribution was experimentally evaluated for a hundred random corrections. 
\textcolor{modif}{ 
As expected, the error increases with the number of switching actuators, as illustrated by Fig. \ref{fig:linerror_validation} showing the approximation error in function of the number of switching actuators. Also, the error is found to be uncorrelated with the following factors: distance to base point where influence vectors are evaluated, number of already ON actuators and motion directions. Furthermore, the contribution to the error of each actuator shows no signifiant variations from one to another. It was therefore possible to characterize the approximation error as a random variable function of the number of switching actuators only.} 
The standard deviation of the error $\boldsymbol{\sigma}_\epsilon$ on each degree of freedom is computed from the experimental sample, and found to be about $0.35\sqrt{s_n}$ mm on $t_x$ and $t_y$ and $0.025\sqrt{s_n}$ degrees on $R_x$ and $R_y$, where $s_n$ is the number of switching actuators.

%The error distribution does not appear to be correlated with any other variable. Error trends in function of the states $\boldsymbol{x}$ or actuators inputs $\boldsymbol{u}$ could have been compensated with a feedback linearization loop $\boldsymbol{\tilde{d}}_k = f(\boldsymbol{u},\boldsymbol{x})$. 

\subsubsection{Cost Function}
\label{sec:cost}

The cost function defines the value of a given switch vector $\boldsymbol{b}$ for a given situation, i.e.\textcolor{modif}{,} an error vector $\boldsymbol{x}_e$ and available influence vectors $\boldsymbol{\tilde{d}}_k$. Three cost functions were evaluated: the norm of the resolution error $\left\| \boldsymbol{\epsilon}_r \right\|$, the inverse of the on-target probability $1/P_{target}$ and the norm of the resolution error plus a penalty proportional to the number of switching actuators $\left\| \boldsymbol{\epsilon}_r \right\|+ 0.2s_n$. Moreover, independant weight for each DOF can be set. For instance, if it is desired to achieve the best possible in-plane translation $\boldsymbol{t}$ accuracy without any consideration for the needle-guide orientation $\boldsymbol{R}$, then both $R_x$ and $R_y$ weights can be set to zero. Such cases are referred as translation-only targets. The 3 cost functions are experimentally evaluated on the prototype for 3 static translation-only targets. TABLE \ref{tab:costfunction} shows the mean results of 3 tests per target, $\epsilon$ is the final absolute translation error and $n_f$ is the number of iterations used. A significant performance improvement is founds for the two cost functions penalizing the number of switching actuators. However, no difference is found between the probabilistic cost function and the simplified proportional version. Hence, it is the simplified cost function that is used in the remaining experiments since its evaluation is faster than the evaluation of the probabilistic cost function.

\begin{table}[htb]
	\centering
		\begin{tabular}{ c  c  c  c  c  c  c }
		\hline
			 & \multicolumn{2}{c}{Target \#1} & \multicolumn{2}{c}{Target \#2} & \multicolumn{2}{c}{Target \#3} \\ \hline
			Cost Function &$n_f$&$\epsilon$&$n_f$&$\epsilon$&$n_f$&$\epsilon$\\ \hline\hline
			$\left\| \boldsymbol{\epsilon}_r \right\|$       & 5 & 0.5 mm & 4 & 0.9 mm & 4 & 0.9 mm \\
			$1/P_{target}$  & 3 & 0.5 mm & 1 & 0.3 mm & 3 & 0.4 mm \\
			$\left\| \boldsymbol{\epsilon}_r \right\|+ 0.2s_n$   & 3 & 0.5 mm & 1 & 0.3 mm & 3 & 0.4 mm \\
			\hline
		\end{tabular}
	\caption{Cost functions experimental evaluation}
	\label{tab:costfunction}
	\vspace*{-15pt}
\end{table}

\subsubsection{Optimization Algorithm} 
\begin{figure}[htbp]
	\centering
		\includegraphics[width=0.45\textwidth]{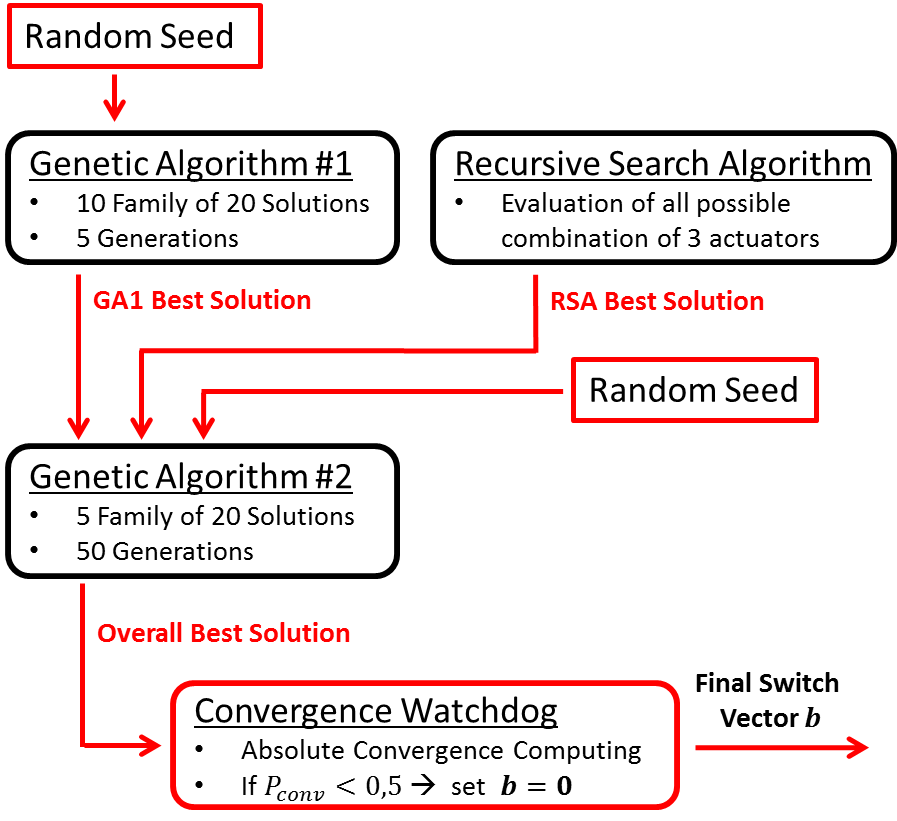}
	\caption{Implemented optimization algorithm.}
	\label{fig:algo}
\end{figure}

The optimization algorithm that finds a switch vector $\boldsymbol{b}$ minimi\textcolor{modif}{z}ing the cost function during each iteration loop is shown in Fig. \ref{fig:algo}. The algorithm optimizes a solution in about 0.3 sec, which is acceptable since the bandwidth bottleneck of the static controller is a three seconds delay imposed between each iterations to let the robot reach a steady state between each correction. 

%Evaluating all possible $2^{20}$ corrections, i.e.\textcolor{modif}{,} possible switch vectors $\boldsymbol{b}$, takes around 100 sec and is impracticable. 

Since a good solution has a low number of switching actuators, all solutions that switch only three or less actuators are evaluated first. In parallel, a widespread, i.e.\textcolor{modif}{,} many members and few generations, genetic algorithm optimizes a solution. The best solutions of those two searches are then inserted in the population of another genetic algorithm with fewer members but many generations that complete the optimization more locally.  The random seeds of both genetic algorithm are randomly generated binary switch vectors $\boldsymbol{b}$ with four switching actuators or more. \textcolor{modif}{For the selection step, the \textit{roulette-wheel selection} method is used. For the crossing step, segments of 5 bits are randomly exchanged between solutions, with a probability of 0.1 for each bit to be the base of a crossing. Then, for the mutation step, a switching probability of 0.025 is used for each bit. The algorithm performances are however subject to improvement as the focus of this work was not put on the optimization of the algorithmic aspect.}

%A particular fact of this algorithm that should be noted is that for a given situation, i.e.\textcolor{modif}{,} an error vector $\boldsymbol{x}_e$ and available influence vectors $\boldsymbol{\tilde{d}}_k$, the outcome can be different each time because of the randomness involved with the seeds.

\subsection{Static Controller Results}

\begin{figure}[htb]
	\centering
		\includegraphics[width=0.50\textwidth]{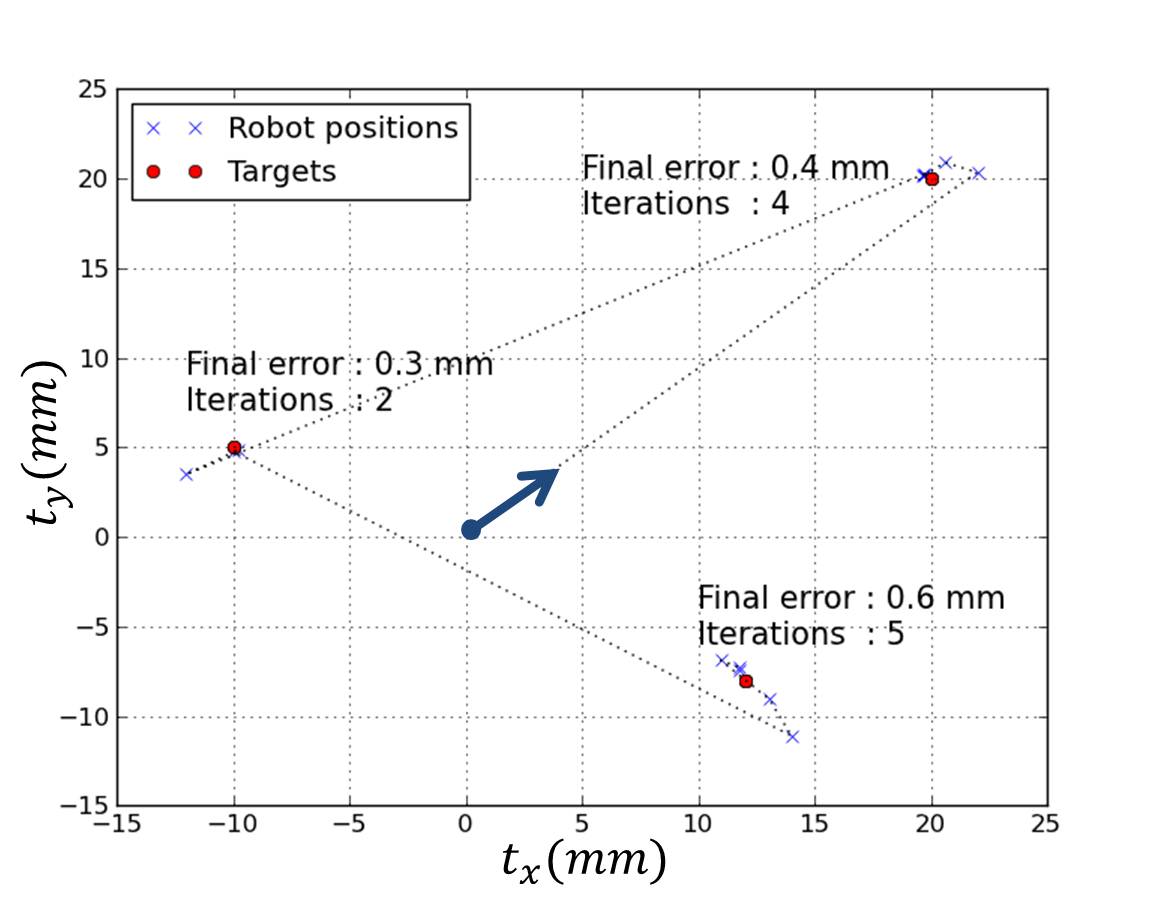}
	\caption{Static control results for translation-only targets.}
	\label{fig:static}
\end{figure}

\begin{figure}[htb]
	\centering
		\includegraphics[width=0.45\textwidth]{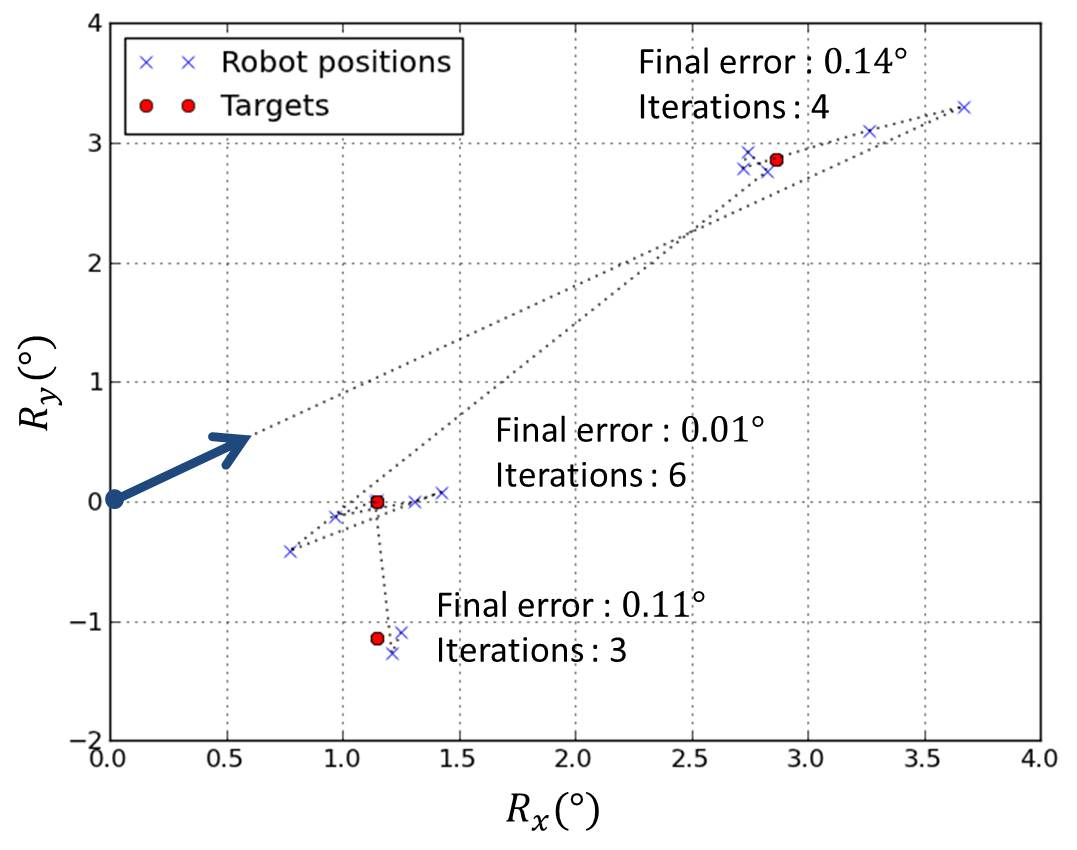}
	\caption{Static control results for rotation-only targets.}
	\label{fig:staticR}
\end{figure}

\begin{figure}[htb]
        \centering
        \subfloat[Translation]{
        \includegraphics[width=0.22\textwidth]{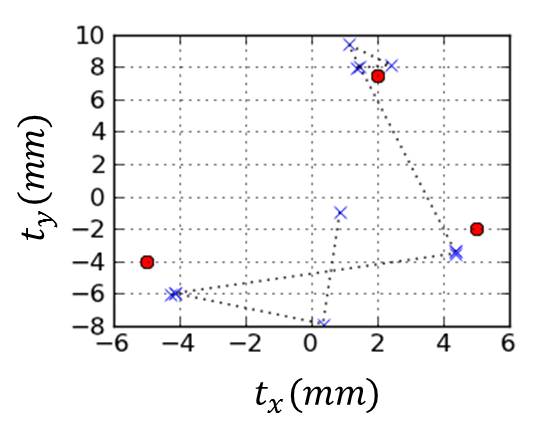}}
        \subfloat[Rotation]{
        \includegraphics[width=0.22\textwidth]{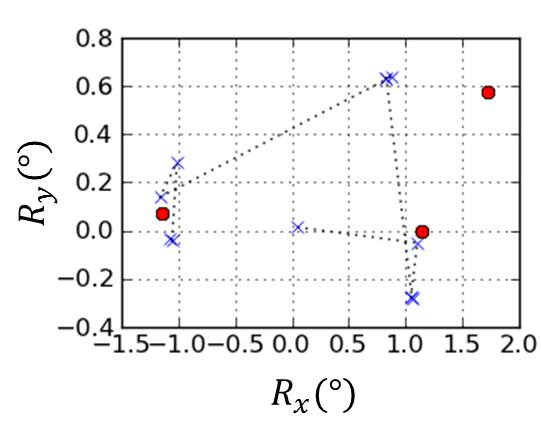}}
        \caption{Static control results for simultaneous control of translation and rotation.}\label{fig:staticTR}
\end{figure}

\textcolor{modif}{To evaluate the static controller\textcolor{modif}{, t}he same three randomly-selected targets are used for all static controller experiments. Also, a three-second delay is imposed to let the robot stabilize to a steady state between each correction.} Typical performance of the static controller is shown on Fig. \ref{fig:static} for translation-only targets, on Fig. \ref{fig:staticR} for rotation-only targets and on Fig. \ref{fig:staticTR} for simultaneous translation and rotation targets.

The static controller performs well for controlling two DOFs while letting two free. Sub-millimeter accuracy, almost at the resolution limit of the cameras, is reached with on average 4 iterations for translation-only targets. The accuracy is lower, in the order of 1-2 mm, see Fig. \ref{fig:staticTR}, for the simultaneous control of the four DOFs. This limited performance is explained by the limited discrete workspace of the 20 actuators binary robots. Ten bits are available per DOF which means 1024 discrete positions per DOF when two DOFs are controlled. Five bits are available per DOF which means only 32 discrete positions per DOF when all four DOFs are controlled. This problem can be addressed by using systems with more binary actuators. Adding only a few actuators would drastically improve the resolution since the relation is exponential. The rest of this paper presents experiments with translation-only targets.

\begin{figure}[htbp]
	\centering
		\includegraphics[width=0.50\textwidth]{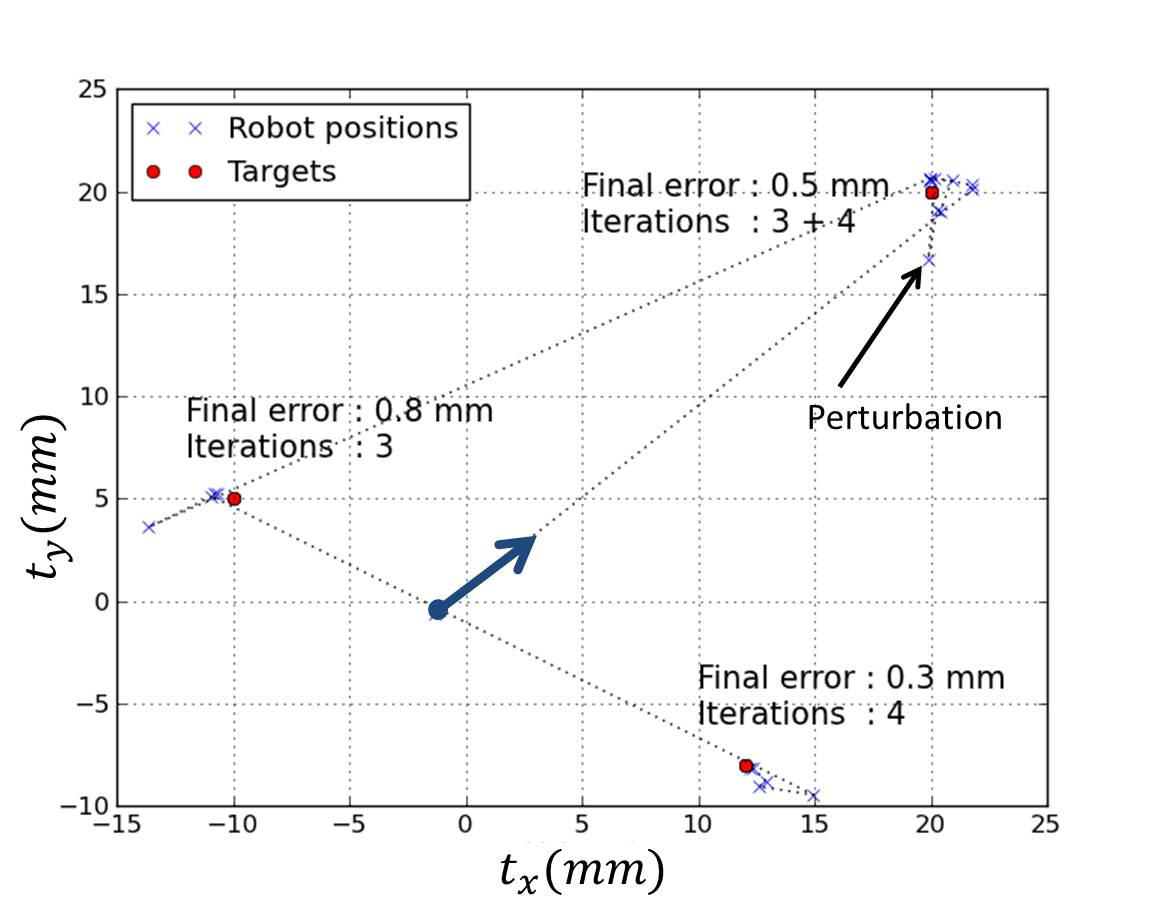}
	\caption{Static control, with a perturbation applied after three iterations.}
	%\vspace*{-5pt}
	\label{fig:staticp}
	%\vspace*{-5pt}
\end{figure}

Fig. \ref{fig:staticp} show the controller's ability to reject perturbations caused by a load that was attached to the robot after the third iteration. The controller is able to correct the perturbation-induced error and system performance is not affected for the following targets even though the load is still on the robot.

\begin{figure}[htb]
	\centering
		\includegraphics[width=0.50\textwidth]{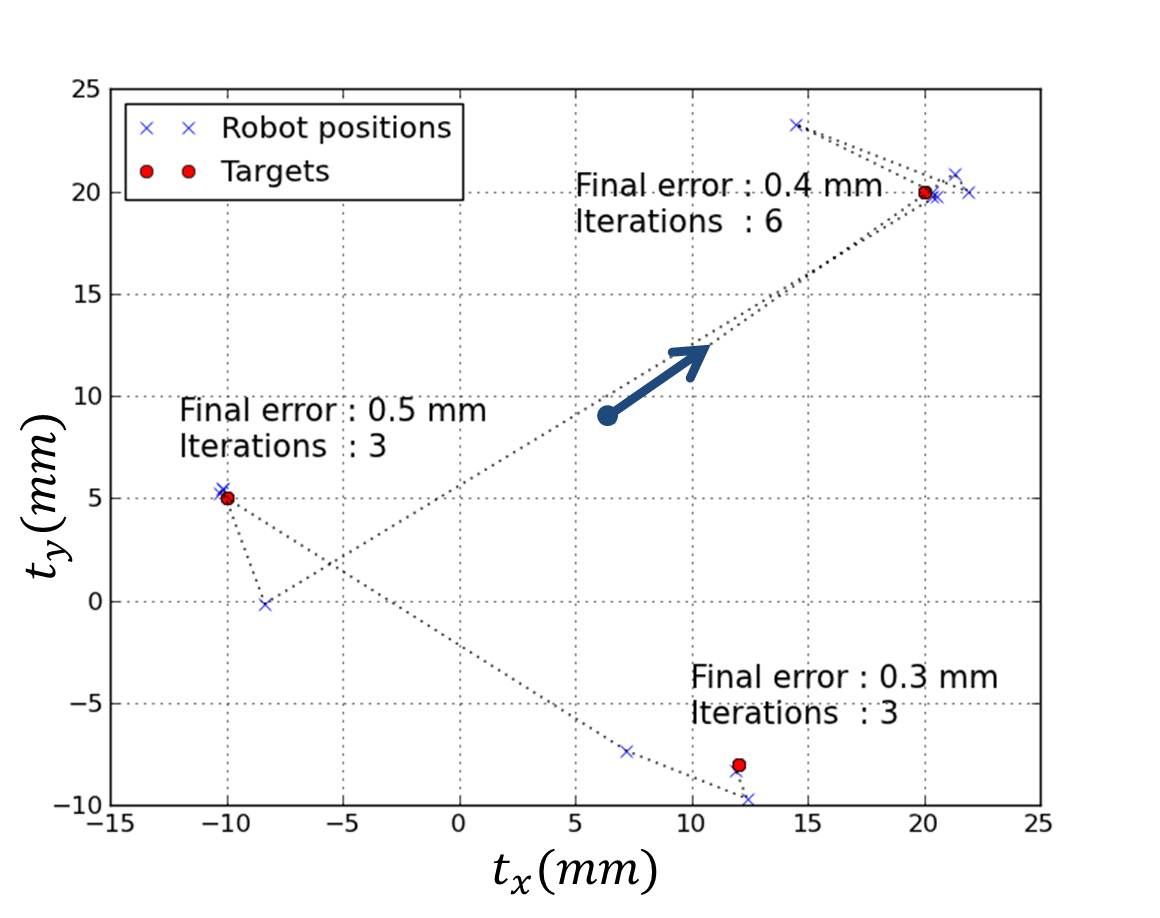}
	\caption{Static control, with three actuators failures undetected.}
	\vspace*{-5pt}
	\label{fig:staticf}
	\vspace*{-5pt}
\end{figure}

\begin{figure}[htb]
	\centering
		\includegraphics[width=0.50\textwidth]{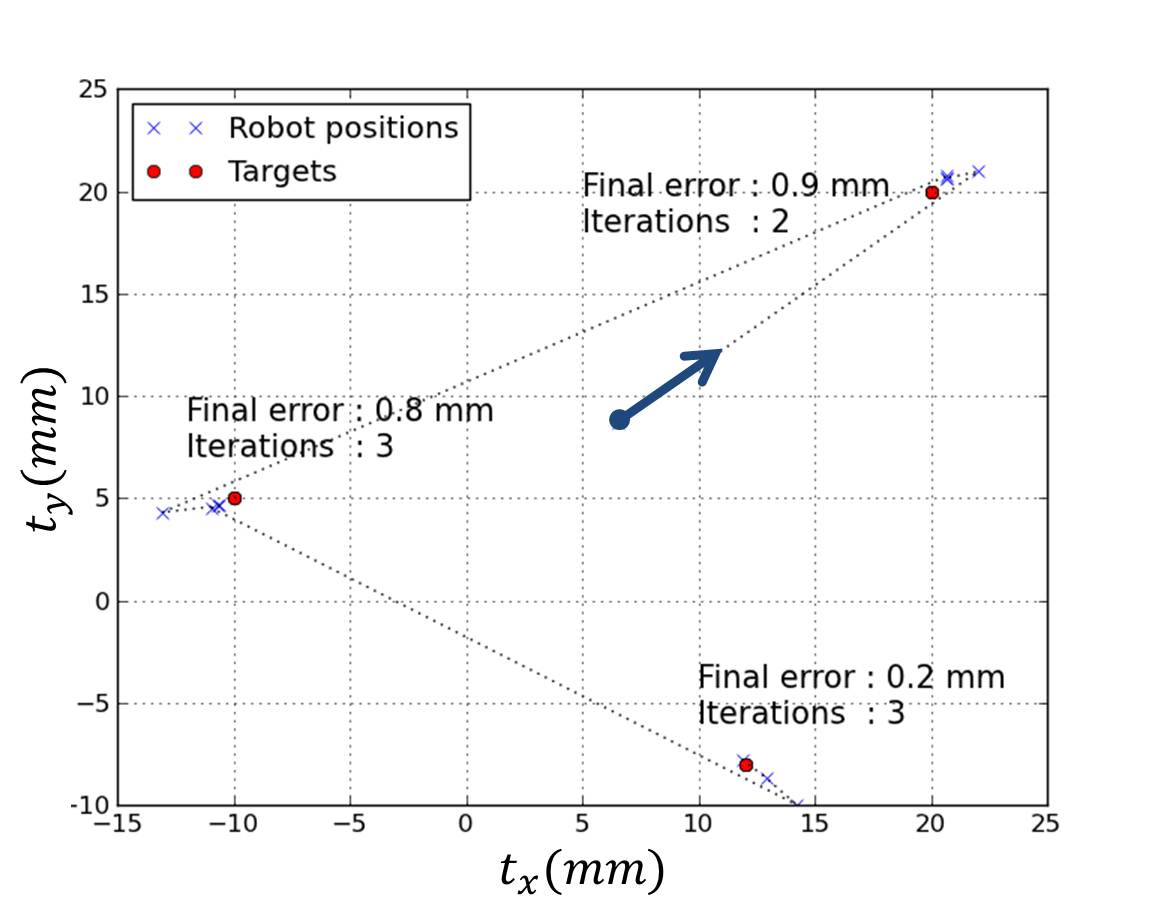}
	\caption{Static control, with three actuators failures detected.}
	\vspace*{-5pt}
	\label{fig:staticfd}
	\vspace*{-5pt}
\end{figure}

Fig. \ref{fig:staticf} shows the ability of the controller to operate in the event of actuator failures. In this experiment, three valves were intentionally blocked open. Fig. \ref{fig:staticfd} simulates the case where these failures could be detected and accounted for by removing defective actuators from the list of possible actuators to recruit.
The robot makes inaccurate moves each time the computed correction uses those actuators for the case with undetected failed actuators, but these errors are corrected by following iterations. Erratic moves stop as soon as failures are detected. However, system resolution diminishes for each out-of-service actuators. Table \ref{tab:staticperf} shows the results of the controller performances for up to 10 actuator failures out of 20. Since the controller is not determinist (random seeds), corrections chosen and performance may be different for two identical situations. Moreover, for some rare lucky situations it could happen that the unpredictable approximation error $\epsilon_a$ cancel out the resolution error $\epsilon_r$. This is why the presented results should be used with care as the randomness involved can bring odd results, such as the better performance of the controller with 10 actuator failures than none, for target \# 1. 

\begin{table}[htb]
	\centering
		\begin{tabular}{ c  c  c  c  c  c  c }
		\hline
			 & \multicolumn{2}{c}{Target \#1} & \multicolumn{2}{c}{Target \#2} & \multicolumn{2}{c}{Target \#3} \\ \hline
			Failures &$n_f$&$\epsilon$&$n_f$&$\epsilon$&$n_f$&$\epsilon$\\ \hline\hline
			None        & 4 & 0.4 mm & 2 & 0.3 mm & 5 & 0.6 mm \\
			3           & 6 & 0.4 mm & 3 & 0.5 mm & 3 & 0.3 mm \\
			3 detected  & 2 & 0.9 mm & 3 & 0.8 mm & 3 & 0.2 mm \\
			6           & 4 & 0.9 mm & 9 & 0.7 mm & 4 & 0.4 mm \\
			6 detected  & 3 & 0.7 mm & 3 & 0.6 mm & 2 & 0.3 mm \\
			10          & 2 & 0.3 mm & 9 & 6.5 mm & 9 & 1.8 mm \\
			10 detected & 1 & 0.2 mm & 2 & 1.4 mm & 2 & 1.7 mm \\
			\hline
		\end{tabular}
	\caption{Static controller performance with actuators failures}
	\label{tab:staticperf}
	\vspace*{-15pt}
\end{table}

\subsection{Dynamic Controller Results}

\begin{figure}[htb]
	\centering
		\includegraphics[width=0.50\textwidth]{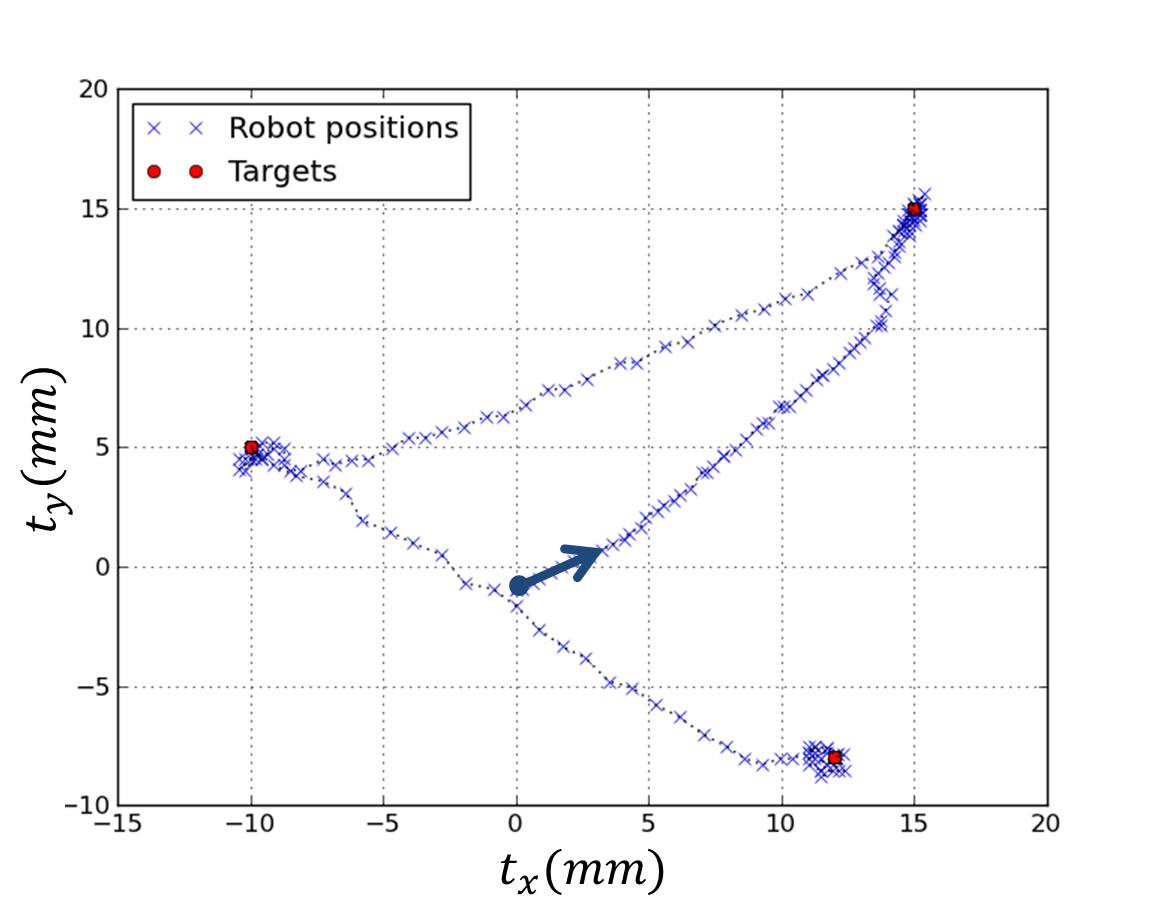}
	\caption{Dynamic controller with static targets.}
	%\vspace*{-5pt}
	\label{fig:dynamic}
	%\vspace*{-5pt}
\end{figure}

\begin{figure}[htb]
	\centering
		\includegraphics[width=0.50\textwidth]{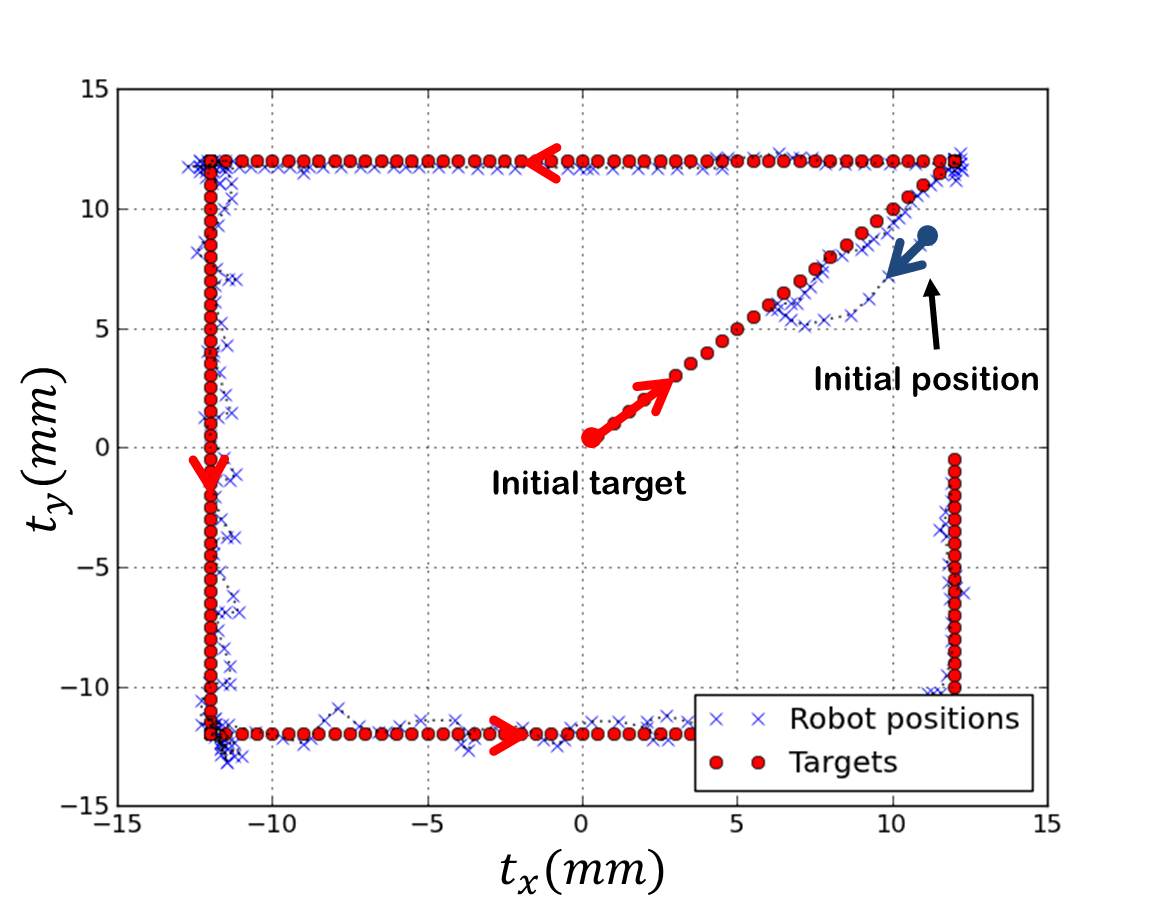}
	\caption{Dynamic controller with a tracking task.}
	%\vspace*{-5pt}
	\label{fig:tracking}
	%\vspace*{-5pt}
\end{figure}

\begin{figure}[htb]
	\centering
		\includegraphics[width=0.45\textwidth]{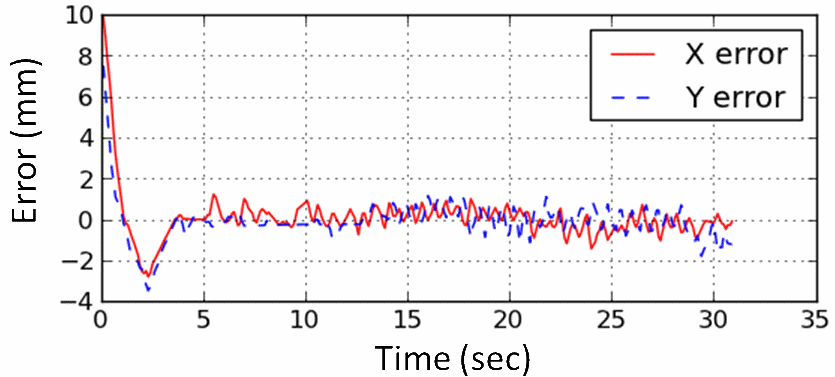}
	\caption{Tracking error of the dynamic controller.}
	%\vspace*{-5pt}
	\label{fig:trackinge}
	%\vspace*{-5pt}
\end{figure}

The experimental set-up was not designed for dynamic experiments and was limited to a 5 Hz control bandwidth because of cameras-induced delays. Such limitations are particularly problematic for sliding mode controllers since the chattering is proportional to control loop delays \cite{perruquetti_sliding_2002}. Fig. \ref{fig:dynamic} shows results with static targets and Fig. \ref{fig:tracking} and \ref{fig:trackinge} show results for a trajectory following task. Results show good potential considering the low control loop frequency. Inevitable chattering occurred, with an oscillation amplitude of about 1 mm.

\begin{figure}[htb]
	\centering
		\includegraphics[width=0.50\textwidth]{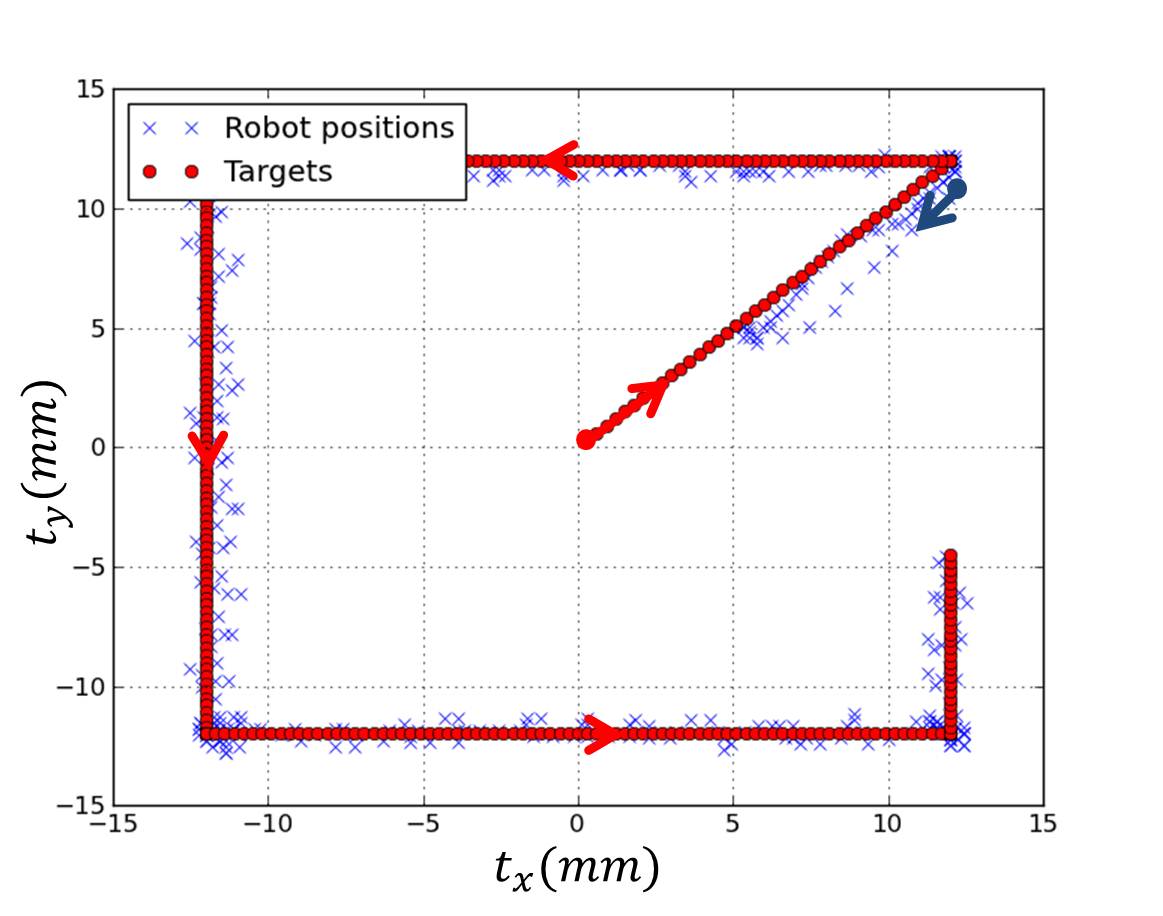}
	\caption{Dynamic controller with a tracking task with 4 actuators faults.}
	%\vspace*{-5pt}
	\label{fig:tracking4}
	%\vspace*{-5pt}
\end{figure}

\begin{figure}[htp]
	\centering
		\includegraphics[width=0.50\textwidth]{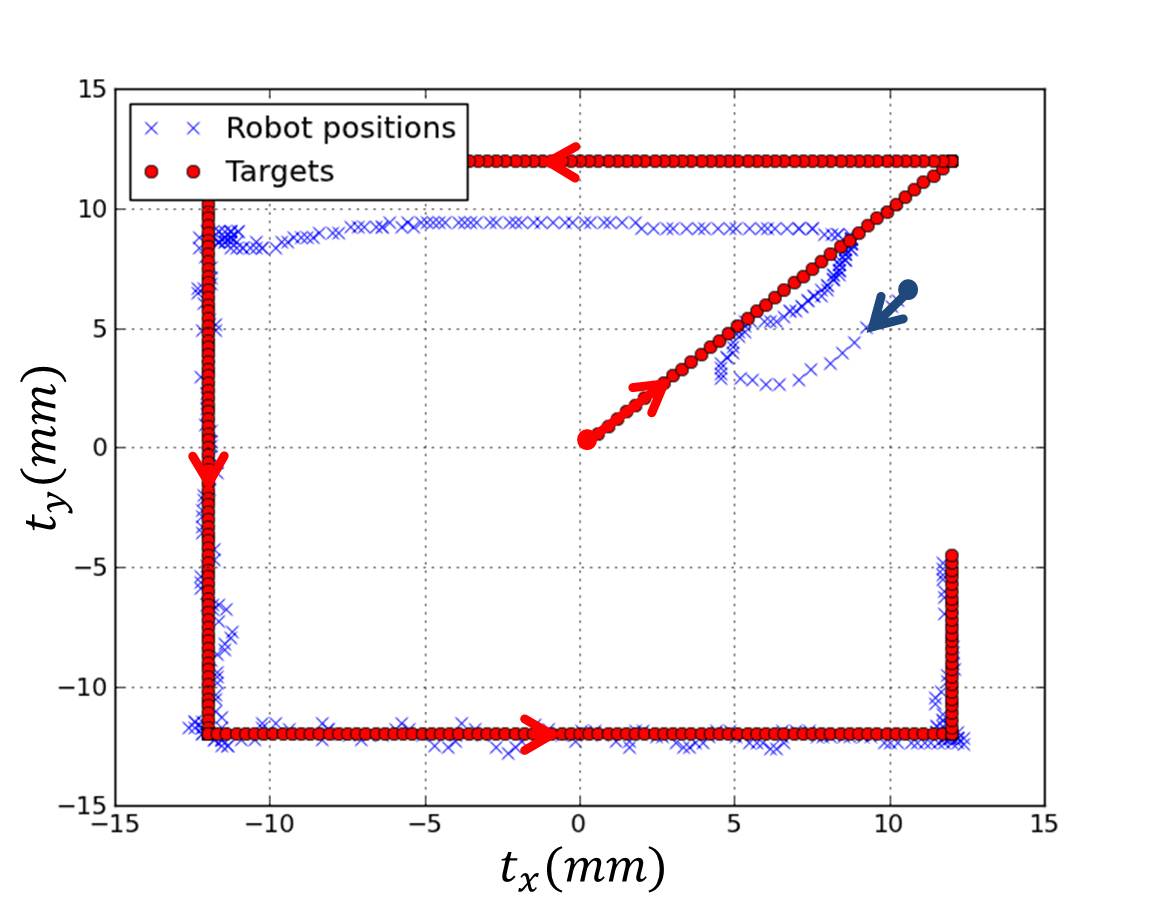}
	\caption{Dynamic controller with a tracking task with 7 actuators faults.}
	%\vspace*{-5pt}
	\label{fig:tracking7}
	%\vspace*{-5pt}
\end{figure}

\begin{figure}[htp]
	\centering
		\includegraphics[width=0.50\textwidth]{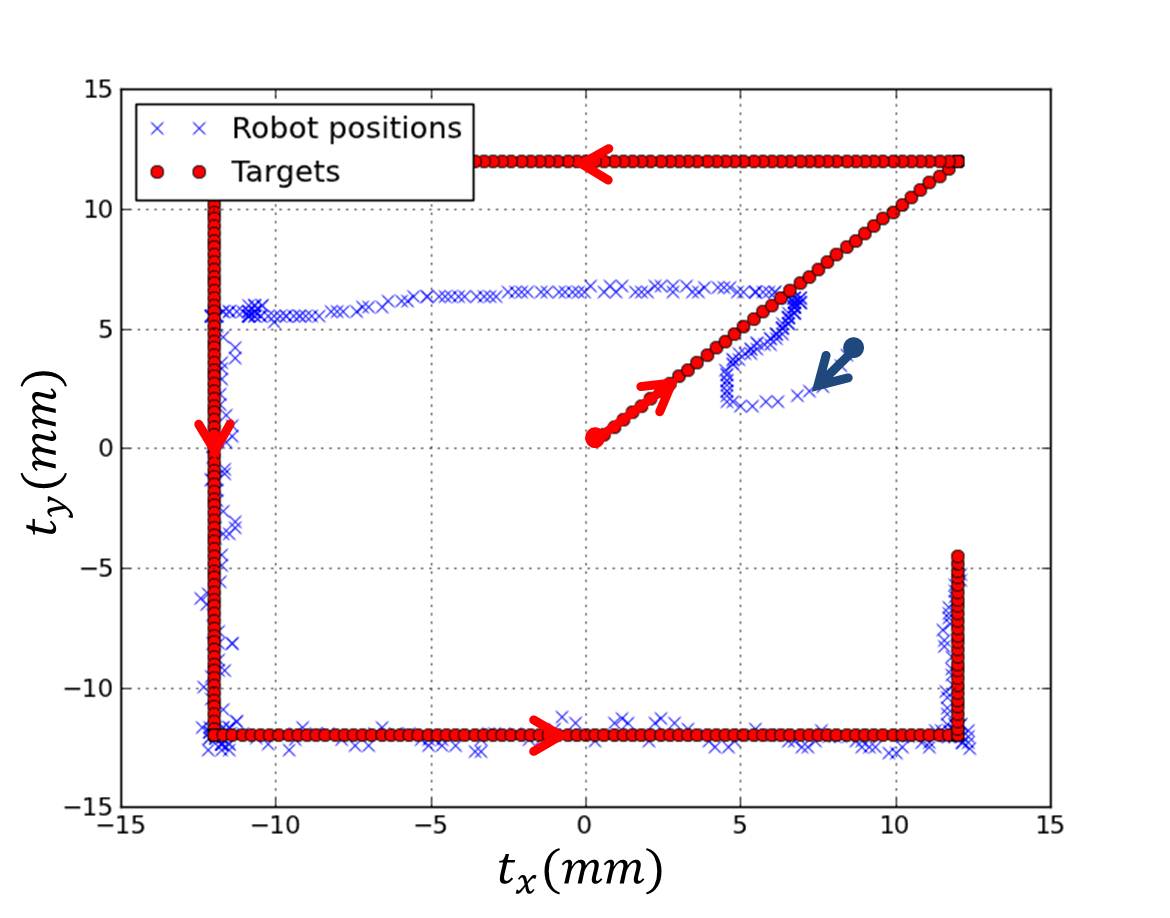}
	\caption{Dynamic controller with a tracking task with 10 actuators faults.}
	%\vspace*{-5pt}
	\label{fig:tracking10}
	%\vspace*{-5pt}
\end{figure}

\begin{figure}[htp]
	\centering
		\includegraphics[width=0.50\textwidth]{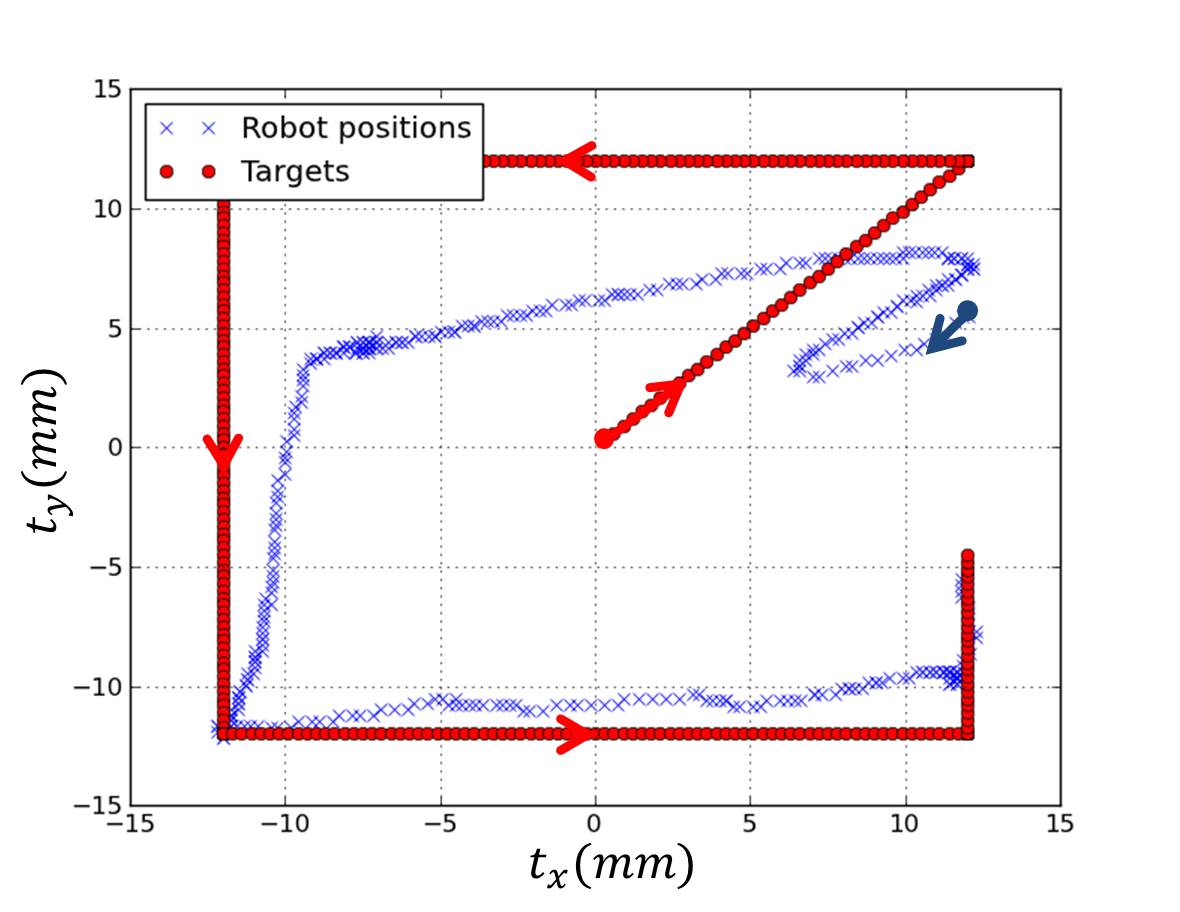}
	\caption{Dynamic controller with a tracking task with 14 actuators faults.}
	%\vspace*{-5pt}
	\label{fig:tracking14}
	%\vspace*{-5pt}
\end{figure}

Fig. \ref{fig:tracking4}, Fig. \ref{fig:tracking7}, Fig. \ref{fig:tracking10} and Fig. \ref{fig:tracking14} present tracking results for respectively 4, 7, 10 and 14 blocked valves out of 20 to simulate actuators failures. The proposed bang-bang controller is found to be very robust to actuator failures. In fact, since the control decision is independent for each actuator, the controller itself is insensitive to broken actuators, although failures will physically limit the system. For instance, with 7 and 10 out-of-service actuators (see Fig. \ref{fig:tracking7} and Fig. \ref{fig:tracking10}), the robot was still able to track trajectories but could not reach the upper part of the workspace, as too many actuators pushing up were unavailable.

\section{CONCLUSION AND \textcolor{modif}{OUTLOOK}}

In this paper, a closed-loop control scheme that can be used on any binary actuated system that is naturally stable is proposed. No analytical model of the system is needed since the scheme relies on experimentally identified influence vectors. A static iterative controller adresses point-to-point tasks and a dynamic controller adresses motion control tasks. The static controller uses a genetic algorithm that searches for optimal combinations of actuators to recruit, iteratively, in order to correct the system error. Corrections are evaluated using the influences vectors and the superposition principle. Experimental results show the static controller to be effective and robust against perturbations and actuators failures. The dynamic controller determines independently the ON-OFF state of each actuators by comparing the direction of their influence vectors with the direction of \textcolor{modif}{a composite} error vector. Experimental results show high potenial, i.e.\textcolor{modif}{,} motion tracking with robustness to perturbations and actuators failures, despite the limited test bench bandwidth. Future work should however include tests using a real-time set-up to evaluate the full potential of the proposed dynamic controller.

The proposed controllers can be used as a base framework for many future developments of binary or cellular-like robots. For instance, an adaptive scheme could be incorporated where the influence vectors could be identified and calibrated online\textcolor{modif}{,} as well as non-linear behaviors and actuator failures. The influence vectors approach could also be expanded to non-binary robots by adding adjustable amplitudes to the influence vectors. Moreover, it would be interesting to develop a hybrid controller that could switch from the dynamic controller to the static controller to limit control activity when the targets become static. At last, a positive non-zero threshold (see eq. \eqref{eq:dynamiccontroller}) for the proposed bang-bang controller, that could be stochastic for large numbers of actuators\cite{ueda_broadcast-probability_2006}, could help moderate the chattering behavior.

%%%%%%%%%%%%%%%%%%%%%%%%%%%%%%%%%%%%%%%%%%%%%%%%%%%%%%%%%%%%%%%%%%%%%%%%%%%%%

\section*{ACKNOWLEDGMENTS}
Special thanks to Genevi\`{e}ve Miron for the robot development, Marc Denninger for the 3D graphic art and Benoit Heintz for the electronics of the experimental set-up.

\bibliographystyle{IEEEtran}
%\bibliography{binarycontrolBIB2,binarycontrolBIB}
\bibliography{main}

%\newpage


%\vfill



%%%%%%%%%%%%%%%%%%%%%%%%%%%%%%%%%%%%%%%%%%%%%%%%%%%%%%%%%%%%%%%%%%%%%%%%%%%%%

%************************************************************************
% References                                                                     
%************************************************************************

\bibliographystyle{IEEEtran}
\bibliography{main}

\end{document}